\documentclass[12pt]{article}
\usepackage[utf8]{inputenc}
\usepackage{cite}
\usepackage{amsfonts}
\usepackage{amsmath}
\usepackage{amsthm}
\usepackage{enumitem}
\usepackage{amssymb}
\usepackage{caption}
\usepackage{graphicx}
\usepackage{mathrsfs}
\usepackage{multirow}
\usepackage{xcolor}
\usepackage{setspace}
\usepackage{booktabs}
\usepackage{latexsym, mathdots, array, extarrows}
\usepackage{algpseudocode,float}
\usepackage{lipsum}
\usepackage{subfig}
\usepackage{pifont}
\usepackage{diagbox}

\usepackage{amsmath,amsfonts,amsthm,amssymb}
\usepackage{algorithm2e}
\usepackage{bm,dsfont}
\usepackage{booktabs}
\usepackage{graphicx}
\usepackage{array}
\usepackage{xcolor}
\usepackage{sidecap}
\usepackage{listings}
\usepackage{times}
\usepackage{url}
\usepackage{hyperref}

\usepackage{cleveref}

\usepackage{geometry}
\newtheorem{theorem}{Theorem}

\newtheorem{lemma}{Lemma}
\newtheorem{assumption}{Assumption}

\newcommand{\ba}{\bm{a}}
\newcommand{\bc}{\bm{c}}
\newcommand{\bG}{\bm{G}}
\newcommand{\bD}{\bm{D}}

\newcommand{\bs}{\bm{s}}
\newcommand{\bx}{\bm{x}}

\allowdisplaybreaks[4]

\title{On the Convergence of (Stochastic) Gradient Descent for Kolmogorov--Arnold Networks
}

\author{Yihang Gao
\thanks{Department of Mathematics, National University of Singapore. Email: gaoyh@nus.edu.sg}\and Vincent Y. F.  Tan\thanks{Department of Mathematics and Department of Electrical and Computer Engineering, National University of Singapore. Email: vtan@nus.edu.sg}}

\date{}

\begin{document}

\maketitle
\begin{abstract}
Kolmogorov--Arnold Networks (KANs), a recently proposed neural network architecture, have gained significant attention in the deep learning community, due to their potential as a viable alternative to multi-layer perceptrons (MLPs) and their broad applicability to various scientific tasks. 
Empirical investigations demonstrate that KANs optimized via stochastic gradient descent (SGD) are capable of achieving near-zero training loss in various machine learning (e.g., regression, classification, and time series forecasting, etc.) and scientific tasks (e.g., solving partial differential equations). 
In this paper, we provide a theoretical explanation for the empirical success by conducting a rigorous convergence analysis of gradient descent (GD) and SGD for two-layer KANs in solving both regression and physics-informed tasks. 
For regression problems, we establish using the neural tangent kernel perspective that GD achieves global linear convergence of the objective function when the hidden dimension of KANs is sufficiently large. We further extend these results to SGD, demonstrating a similar global convergence in expectation.
Additionally, we analyze the global convergence of GD and SGD for physics-informed KANs, which unveils additional challenges due to the more complex loss structure.
This is the first work establishing the global convergence guarantees for GD and SGD   applied to optimize KANs and physics-informed KANs.
\end{abstract}

\section{Introduction}
The Kolmogorov--Arnold Representation Theorem (KART)~\cite{kolmogorov1961representation,kolmogorov1957representation,braun2009constructive} states that any continuous (multivariate) function $f: \mathbb{R}^{n} \to \mathbb{R}$ can be decomposed into a sum of univariate continuous functions. Specifically, there exists a set of univariate functions $\{\phi_{p,q}\}_{p=1,q=1}^{n, 2n+1}$ and $\{\Phi_{q}\}_{q=1}^{2n+1}$ such that
\begin{equation}
\label{eq_KART}
    f(\bm{x}) = \sum_{q=1}^{2n+1} \Phi_{q}\left(\sum_{p=1}^{n} \phi_{p,q}\left( x_{p}\right)\right),
\end{equation}
where $\bx = [x_1, x_2, \ldots, x_n]^{\top}$. 
Inspired by this theorem, efforts have been made to develop novel network architectures, as an alternative to multi-layer perceptrons (MLPs) supported by the Universal Approximability Theory (UAT).  
However, previous approaches have been largely restricted by the limited width of networks, due to the representation in \Cref{eq_KART}, where the hidden dimension is constrained to be twice the input dimension.
Moreover, the implicit representation function $\phi_{p,q}$ in KART can be extremely non-smooth and complicated, making it challenging to approximate by splines or polynomials in practice. 
These problems limit the practical applicability of KART-based networks and account for the failure of the previous network design, which focuses mostly on enhancing the network capabilities by directly approximating these implicit representation functions. 
It has been decades since researchers began to consider networks based on KART. 
However, in a recent breakthrough, inspired by the empirical success of deep and wide MLPs,  Liu et al.~\cite{liu2024kan} successfully generalized the Kolmogorov--Arnold Network (KAN) to be multi-layer and arbitrarily wide. The new design overcomes the limitations of traditional KART-based networks and demonstrates superior empirical performance, particularly in symbolic formula representation and scientific tasks.

Several studies have shown that KANs perform comparably or even outperform MLPs, highlighting the potential as an alternative to MLPs, on a variety of tasks, including computer vision, reinforcement learning, and physics-informed machine learning. 
For example, Bodner et al.~\cite{bodner2024convolutional} and Drokin~\cite{drokin2024kolmogorov} independently designed convolutional KANs as an alternative to standard CNNs for computer vision tasks. They further extended these architectures to construct a U-Net-like structure using convolutional KANs, achieving state-of-the-art performances~\cite{li2024u}. 
In the domain of time-series forecasting, Genet and Inzirillo~\cite{genet2024tkan} introduced Temporal KANs by substituting MLPs in RNNs with KANs, thereby enhancing both accuracy and efficiency in capturing complex temporal patterns~\cite{genet2024temporal,xu2024kolmogorov,han2024kan4tsf}.
KAN-based reinforcement learning algorithms have also been explored in~\cite{kich2024kolmogorov,guo2024kan}, demonstrating comparable or superior performance to MLPs, often with fewer parameters. 
Additionally, Zhang and Zhang~\cite{zhang2024graphkan} used KANs as Graph Neural Networks to extract features and capture dependencies within graphs. 
Efforts to incorporate KANs into Transformers validate their potential in scaling up large language and vision models. They attempted to address the challenges of the incorporation of KANs, such as the non-parallel nature, instability of base functions, inefficient forward computation, and difficulties in weight initialization.
KANs have also shown promise in scientific tasks. Wang et al.~\cite{wang2024kolmogorov} conducted comprehensive and convincing empirical studies on physics-informed KANs for solving partial differential equations (PDEs), including strong form, energy form, and inverse of PDEs. Moreover, Aghaei~\cite{aghaei2024kantrol} and Toscano et al.~\cite{toscano2024inferring} applied physics-informed KANs to concrete physical problems. Their results demonstrate that KANs significantly outperform MLPs regarding the accuracy and convergence speed for solving PDEs, highlighting the potential of physics-informed KANs in AI for solving complex physical systems~\cite{abueidda2024deepokan}. More investigations and surveys of KANs can be found in~\cite{rigas2024adaptive,howard2024finite,nagai2024kolmogorov,cheon2024demonstrating,knottenbelt2024coxkan,qiu2024relu,polar2021deep,chen2024gaussian,hou2024comprehensive,yu2024kan,liu2024kan2}.

Notably, the empirical results~\cite{wang2024kolmogorov,liu2024kan,ss2024chebyshev,aghaei2024rkan} indicate that the KANs optimized by SGD and its variants (e.g., Adam~\cite{kingma2015adam}) can achieve very low (near-zero) training loss across various tasks, although the objective functions are extremely nonlinear and nonconvex. This observation raises a natural question:  \textit{How exactly does the SGD optimize KANs and achieve near-zero training loss?}

Existing analysis of the gradient descent (GD) usually depends on smoothness~\cite{carmon2018accelerated, li2019convergence}, Lipschitzness~\cite{carmon2018accelerated, nesterov2006cubic}, and convexity~\cite{duchi2011adaptive, kingma2015adam, reddi2019convergence} to ensure global convergence. Consequently, these works are not directly applicable to deep learning whose loss functions are highly non-convex and not necessarily smooth. 
While much of the literature focuses on convergence to local optima, numerical experiments have shown that both GD and SGD nearly achieves  {\em global} optima in practice in which the mean-square error approaches zero \cite{zhang2021understanding}. 
This phenomenon in deep learning cannot be explained by classical convergence analysis but is instead attributed to the {\em   overparameterization} of neural networks. Soudry and Carmon~\cite{soudry2016no} demonstrated that all local minima are in fact global minima, for over-parameterized MLPs. Du et al.~\cite{du2018gradient} proved that the GD finds the global optima of over-parameterized MLPs with ReLU activation, for least squares problems. Gao et al.~\cite{gao2023gradient} extended the results for training Physics-Informed Neural Networks (PINNs), utilizing similar techniques. We refer to readers for more related works of GD and SGD~\cite{soltanolkotabi2017learning, xie2017diverse, chizat2018global, jacot2018neural, soltanolkotabi2018theoretical,wang2022and,vaswani2019fast,oymak2020toward,allen2019convergence,xu2024convergence,xu2024convergence2}. 
Despite these advances, the global convergence of SGD for training KANs and physics-informed KANs observed in numerical experiments cannot be fully explained by the aforementioned results.

We aim to address the question by studying two-layer but sufficiently wide KANs on regression and physics-informed problems from neural tangent kernel perspectives. The results can be similarly extended for multi-layer KANs with minor modifications, which we leave as a future investigation. 
Physics-informed tasks, such as finding the numerical solutions of certain important partial differential equations, are among the most significant and promising applications of KANs in scientific tasks, making the investigation of GD and SGD convergence for physics-informed KANs both essential and necessary.
This is the first work establishing the global convergence guarantees for GD and SGD applied to optimize KANs and physics-informed KANs.  Our contributions are summarized as follows:

\noindent (i) We rigorously establish that the vanilla GD achieves global linear convergence of the objective function for regression problems when the hidden dimension of KANs is sufficiently large.

\noindent (ii) We further prove that, when optimized by SGD, the objective function decays exponentially to zero (linear convergence) in expectation.

    \noindent (iii) We analyze the global convergence of GD and SGD for training physics-informed KANs, a   more challenging task in which the objective  function involves the differential operators on input variables. 


\section{Preliminaries}
\label{sec_preliminaries}
In this section, we begin by introducing notations used throughout the paper. We then review the foundational concepts and definitions of KANs, including their mathematical formulation and various extensions. Finally, we discuss the problem setting for applying KANs to regression problems.

\subsection{Notation}
In this paper, we use bold capital letters (e.g., $\bm{A}$) and bold lowercase letters (e.g., $\bm{x}$) to denote matrices and vectors, respectively, while scalars are represented using regular (non-bold) letters (e.g., $a$). The symbol $\lesssim$  is used to denote upper bounds up to a (unimportant) constant factor, i.e., $a \lesssim b$ implies that there exists a positive constant $c$ such that $a \leq c \cdot b$. Similarly, $a \gtrsim b$ indicates $a \geq c\cdot b$ for some positive constant $c$, and $a \simeq b$ is equivalent to $a \lesssim b$ and $b \lesssim a$. We use $[n]$ to denote  $\{1,2,\ldots,n\}$. Additionally, we let $(g \circ f)(x)$ denote the composition $g(f(x))$, and $(g \circ f \cdot h)(x)$ represent the composition combined with multiplication, defined as $(g \circ f \cdot h)(x)=g(f(x)) \cdot h(x)$. The operation $\text{vec}(\cdot)$ reshapes the given matrix or tensor into a column vector. For notational convenience, we simplify $\mathcal{L}(\ba(t), \bc(t))$ as $\mathcal{L}(t)$, if without any ambiguity, which is also applied for other notations such as $\bG$ and $\bs$. 

\subsection{Kolmogorov--Arnold Networks}
According to the primary definition of KANs in \cite{liu2024kan}, the $(\ell+1)$-st layer $\bm{z}_{\ell+1}=[z_{\ell+1,1},\ldots,z_{\ell+1,n_{\ell+1}}]^{\top}$ of a KAN is given by: 
\begin{equation*}
    z_{\ell+1,q} = \sum_{p=1}^{n_{\ell}} \phi_{\ell,p,q}(z_{\ell,p}), \quad q \in [n_{\ell+1}],
\end{equation*}
where $n_{\ell}$ is the width of the $\ell$-th layer and $\phi_{\ell,p,q}$ denotes the $(p,q)$-th representation function in the $\ell$-th layer. In practice, we adopt splines or polynomials to parameterize and approximate $\phi_{\ell,p,q}$ due to their universal approximability. Specifically, we define
\begin{equation*}
    \phi_{\ell,p,q}(z) = \sum_{k=1}^{n_d} a_{\ell,p,q,k} b_{k}(z),
\end{equation*}
where $\{b_{k} \}_{  k \in [n_d]}$ is a set of basis functions.
Sidharth and Gokul~\cite{ss2024chebyshev} suggested using Chebyshev polynomials as basis functions due to their orthogonality, powerful approximation capability, and rapid convergence. 
Aghaei~\cite{aghaei2024rkan,aghaei2024fkan} adopted rational and fractional functions for modeling non-polynomial and more complex behaviors. 
Qiu et al.~\cite{qiu2024relu} replaced the traditional B-splines with ReLU activations of higher-order powers and introduced trainable turning points. This modification significantly enhances GPU acceleration for KANs and mitigates the computational bottlenecks associated with B-splines.
Bozorgasl and Chen~\cite{bozorgasl2024wav} incorporated wavelet functions as basis functions, enabling KANs to efficiently capture both low-frequency and high-frequency components of input sequences. 
Ta et al.~\cite{ta2024fc} explored combinations of radial basis functions, B-splines, wavelet functions, and differences of Gaussians, advocating for future KAN designs to leverage a diverse set of basis functions. 
Seydi~\cite{seydi2024exploring} conducted a comprehensive survey and empirical comparison of various polynomials as basis functions in KANs, contributing to a deeper understanding of their capabilities and limitations.
Notably, the approximability of polynomials, rational, and fractional functions is typically restricted to a bounded domain (e.g., the Chebyshev polynomial is usually defined on $[-1,1]$). Thus, an additional nonlinear transformation is   necessary to map input values into the desired domain~\cite{ss2024chebyshev,aghaei2024rkan,aghaei2024fkan}. Specifically, we can alternatively design:
\begin{equation*}
    \phi_{\ell,p,q}(z) = \sum_{k=1}^{n_d} a_{\ell,p,q,k} b_{k}(\phi(z)),
\end{equation*}
where $\phi$ is a   function that maps the input to a bounded interval. Common and practical choices for $\phi$ include the hyperbolic tangent (tanh) and sigmoid functions. Here, the width $n_{\ell}$ for each layer, the depth $\ell \in [L]$,  the number of basis functions $n_d$, and the selection of basis functions are all considered hyperparameters.

\subsection{Problem Formulation}
We consider a two-layer KAN (i.e., depth $L=2$) with a sufficiently large hidden dimension $m$, formulated as
\begin{equation}
\label{eq_two_layer_kan}
        f(\bm{x};\ba,\bc)
        = \frac{1}{\sqrt{m}}\sum_{q=1}^{m} \sum_{k=1}^{n_d}c_{q,k} b_k\left( \phi\left(\sum_{p_1=1}^{n} \sum_{k_1=1}^{n_d} a_{p_1,q,k_1} b_{k_1}\left(x_{p} \right) \right)\right),
\end{equation}
where $\bx=[x_1,\ldots,x_n]^{\top} \in \mathbb{R}^{n}$ is the input vector. The trainable parameters $\ba \in \mathbb{R}^{m n n_d}$ and $\bc \in \mathbb{R}^{m n_d}$ are defined as:
$\ba=[\ba_1^{\top},\ldots,\ba_{m}^{\top}]^{\top}$ with $\ba_{q}=\text{vec}(\{a_{p,q,k}\}_{p=1,k=1}^{n, n_d})$ and $\bc=[\bc_1^{\top},\ldots,\bc_{m}^{\top}]^{\top}$  with $\bc_{q}=\text{vec}(\{c_{q,k}\}_{k=1}^{n_d})$. Here, $\text{vec}(\cdot)$ represents the vectorization of parameters. 
The coefficients $a_{p,q,k}$ and $c_{q,k}$ are initialized independently as:
\begin{equation}
\label{eq_ntk_initialization}
    a_{p,q,k}, c_{q,k} \sim \mathcal{N}(0,1).
\end{equation}
This initialization follows the NTK-initialization~\cite{du2018gradient}. Our method can be generalized and adapted to other popular initialization schemes (e.g., He's~\cite{he2015delving} and Lecun's~\cite{klambauer2017self} initialization) with only minor modifications to the technical details.

Given a set of training samples $\{(\bx_{i}, y_i)\}_{i=1}^{N}$, the corresponding empirical loss for the two-layer KAN in Equation (\ref{eq_two_layer_kan}) is defined as:
\begin{equation}
        \mathcal{L}\left(\ba,\bc \right)
        \!:=\!  \frac{1}{N} \sum_{i=1}^{N} \left(f_i(\ba,\bc)\! -\! y_i \right)^2
        \!:=\! \sum_{i=1}^{N} s_i(\ba,\bc)^2,\label{eq_loss}
\end{equation}
where $f_i(\ba,\bc)$ denotes $f(\bm{x}_i; \ba,\bc)$ and $s_i(\ba,\bc)$ is the normalized predicted error for i-th sample, defined as $\frac{1}{\sqrt{N}}\left(f_i(\ba,\bc) - y_i\right)$. 
Here, we consider the gradient flow case for clarification, while our main results focus on the more practical GD and SGD algorithms. If the dynamics for training the  parameters follows 
\begin{equation}
\label{eq_gradient_flow}
\begin{split}
        \frac{\mathrm{d} \ba(t)}{\mathrm{d} t} & = -\frac{\partial \mathcal{L} (\ba(t),\bc(t))}{\partial \ba},\\
    \frac{\mathrm{d} \bc(t)}{\mathrm{d} t} & = -\frac{\partial \mathcal{L} (\ba(t),\bc(t))}{\partial \bc},
\end{split}
\end{equation}
then, from the NTK perspective~\cite{du2018gradient}, we obtain
\begin{equation}
\label{eq_dynamic_s}
    \frac{\mathrm{d} \bm{s}(\ba(t),\bc(t))}{\mathrm{d} t} = - \bm{D}^{\top} \bm{D} \bm{s}:= - \bG \bm{s},
\end{equation}
where $\bs = [s_1,\ldots,s_{N}]^{\top}$, and the Gram matrix $\bG$ is defined as
\begin{equation}
\label{eq_gram_matrix}
    \bG = \bm{D}^{\top} \bm{D} \in \mathbb{R}^{N \times N},
\end{equation}
with
\begin{equation*}
    \bm{D} := \left[\begin{array}{ccc}
        \frac{\partial s_1}{\partial \ba} & \ldots & \frac{\partial s_{N}}{\partial \ba} \\
        \frac{\partial s_1}{\partial \bc} & \ldots & \frac{\partial s_{N}}{\partial \bc} 
    \end{array} \right] \in \mathbb{R}^{(m \times n \times n_d + m \times n_d) \times N}.
\end{equation*}
\Cref{eq_dynamic_s} further implies that if the matrix $\bG$ remains strictly positive definite along the trajectory, the prediction error $\bm{s}$ (and thus the empirical loss $\mathcal{L}$) will consistently decay to zero. 
This observation provides valuable insights, as a significant part of our analysis is devoted to demonstrating the positive definiteness of $\bG$ both at initialization and throughout the trajectories of GD and SGD. With these foundational results in place, we are now prepared to delve into the detailed convergence analyses for GD and SGD.

\section{Convergence Analysis for GD}
\label{sec_gd}
In this section, we analyze the convergence of GD when optimizing the loss function defined in \Cref{eq_loss}. Our analysis is built upon the property that the Gram matrix in \Cref{eq_gram_matrix}  remains positive definite along the trajectory of GD. 
The vanilla GD takes the form
\begin{equation}
\label{eq_gd}
\begin{split}
    \ba(t+1) & = \ba(t) -\eta \cdot \frac{\partial \mathcal{L}}{\partial \ba}(\ba(t),\bc(t)),\\
    \bc(t+1) & = \bc(t) -\eta \cdot \frac{\partial \mathcal{L}}{\partial \bc}(\ba(t),\bc(t)),
\end{split}
\end{equation}
where $\eta >0$ is the step size.
Since GD is a (Euler's) discrete version of the continuous gradient flow equation (\Cref{eq_gradient_flow}), it ensures a sufficient decrease in the objective function as long as the Gram matrix is positive definite and the step size is sufficiently small.
We first show that, given a sufficiently large hidden dimension, the Gram matrix at initialization (\Cref{eq_ntk_initialization}) is positive definite with high probability.  Furthermore, under the ``lazy'' training regime of \Cref{eq_gd}, the parameters $\ba(t)$ and $\bc(t)$ remain within the neighborhoods of their initialization $\ba(0)$ and $\bc(0)$, respectively. This, in turn, ensures that the Gram matrix $\bG(t)$ remains positive definite throughout the  whole training process.
By carefully choosing the step size and ensuring a sufficiently large hidden dimension, GD, as an Euler discretization of \Cref{eq_gradient_flow}, approximates the continuous dynamics well and guarantees a sufficient decrease in the objective function $\mathcal{L}$.
The detailed analysis is provided below, and all proofs are presented in \Cref{appendix_proof_gd}.

\begin{assumption}
\label{assump_positive}
    Let $\bG^{\infty}:= \mathbb{E}_{\ba,\bc}\left[\bG\left(\ba,\bc\right) \right]$ denote the expected Gram matrix where the expectation is over the initialization (\Cref{eq_ntk_initialization}).
    We assume that $\bG^{\infty}$ is positive definite, i.e., $\sigma_{\min} := \lambda_{\min}\left(\bG^{\infty}\right)>0$. Moreover, the matrix $\bG^{\infty}$ does not depend on the hidden dimension $m$.
\end{assumption}

The assumption on the positive definiteness of $\bG^{\infty}$ is justified as follows. Note that $\bG$ is the Gram matrix of a tall matrix $\bD \in \mathbb{R}^{(m \times n \times n_d + m \times n_d) \times N}$. Therefore, the positive definiteness of $\bG$ is equivalent to the linear independence of columns of $\bD$, which is likely to hold due to the large row dimension. 
Based on this intuition, we provide a rigorous proof for the positive definiteness of $\bG^{\infty}$ under specific choices of basis and transformation functions. For other variants of KANs with different bases and transformation functions, an extension of the proof may be required.

\begin{lemma}
\label{lemma_positive}
    Assume that the basis functions $\{b_k\}_{k=1}^{n_d}$ are chosen as polynomials of degree less than $n_d$ and the transformation functions as the hyperbolic tangent or sigmoid, then $\sigma_{\min} > 0$ under the condition that all training samples are distinct. Moreover, if no transformation function is applied (i.e., $\phi(z) = z$), then $\sigma_{\min} > 0$ holds if all training samples are linearly independent in the $\Tilde{n}_d$-degree polynomial space, i.e., 
    \begin{equation*}
            \{[x_{i,1},x_{i,1}^{2},\cdots, x_{i,1}^{\Tilde{n}_d}, \cdots, x_{i,n},x_{i,n}^{2},\cdots, x_{i,n}^{\Tilde{n}_d}]\}_{i=1}^{N}
        \end{equation*}
        are linearly independent, where $\Tilde{n}_d = (n_d-1)^2$. 
\end{lemma}

The proof is provided in~\Cref{appendix_proof_lemma_positive}. This lemma suggests that KANs with polynomial basis functions and hyperbolic tangent or sigmoid transformation functions (e.g., Chebyshev KANs~\cite{ss2024chebyshev} and all variants in~\cite{seydi2024exploring}) have an expected Gram matrix $\bG^{\infty}$ that is  positive definite, provided that all training samples are distinct.
Moreover, B-splines basis functions adopted in vanilla KAN can also be regarded as polynomials defined on intervals, where the latter result in the lemma applies. 
For other variants of KANs using rational, fractional, wavelet, or radial basis functions, or their combinations, the proof is more intricate and challenging, which we leave as a topic for future investigation. 
Additionally, we introduce the following mild and realizable assumptions on the basis functions $\{b_{k}\}_{k=1}^{n_d}$ and transformation function $\phi$.

\begin{assumption}
\label{assump_bounded}
    Assume that the transformation function $\phi$ is smooth and uniformly bounded on $\mathbb{R}$, and has  bounded first- and second-order derivatives. Furthermore,  assume that the basis functions $\{b_k\}_{k=1}^{n_d}$ and their derivatives up to the second order are uniformly bounded over the images of $\phi$.
\end{assumption}

Polynomial-related basis functions~\cite{ss2024chebyshev, aghaei2024rkan,aghaei2024fkan}, smooth and bounded transformation functions, such as the hyperbolic tangent and sigmoid, are commonly used to stabilize both the computation and the theoretical expressiveness of the basis functions.  In the vanilla KAN~\cite{liu2024kan}, the transformation function implicitly acts as an identity mapping within the domain of interest, but as a zero mapping outside the domain. This behavior is smooth and bounded almost everywhere, except at specific turning points of spline segments. For radial basis functions~\cite{ta2024fc}, both the basis functions and their derivatives of any order are uniformly bounded on $\mathbb{R}$, thereby satisfying \Cref{assump_bounded}.
The following two lemmas establish properties of the initialized loss function $\mathcal{L}(0)$ and the Gram matrix $\bG(0)$. With high probability, the initialized loss function is bounded, which is consistent with practical observations~\cite{liu2024kan,liu2024kan2,aghaei2024rkan,wang2024kolmogorov}.

\begin{lemma}
\label{lemma_init_loss}
    With probability  at least $1-\delta$ over the initialization, 
$    
        \mathcal{L}(0) \lesssim n_d \cdot \log \left(\frac{N}{\delta}\right).
$     
\end{lemma}

\begin{lemma}
\label{lemma_init_gram}
    Under Assumptions \ref{assump_positive} and \ref{assump_bounded}, and the condition that
    \begin{equation*}
        m \gtrsim \frac{n_d^4 n^2}{\sigma_{\min}^2} \log \left(\frac{N}{\delta}\right) \cdot \left(n_d^2 + \left(\log \frac{m}{\delta}\right)^2\right),
    \end{equation*}
    with probability at least $1-\delta$ over the initialization of $\ba$ and $\bc$, we have $\left\|\bG(0) - \bG^{\infty} \right\|_2 \leq \frac{\sigma_{\min}}{8}$.
\end{lemma}

The  proofs for these lemmas can be found in Appendices~\ref{appendix_proof_lemma_init_loss} and \ref{appendix_proof_lemma_init_gram}. \Cref{lemma_init_gram} indicates that the initialized Gram matrix $\bG(0)$ is close to the expected Gram matrix $\bG^{\infty}$ when the hidden dimension $m$ is sufficiently large. 
To interpret this, consider each element of $\bG(0)$, denoted as $G_{i,j}(0)$, as the average of $m$ independent and identically distributed random variables, i.e., $G_{i,j}(0) = \frac{1}{m} \sum_{q=1}^{m} X_q$. Using concentration inequalities, such as Hoeffding's inequality, we can show that $G_{i,j}(0)$ converges to its expectation $G^{\infty}_{i,j}$ as $m$ becomes large. 
In the next lemma, we identify neighborhoods around the initial parameters $\ba(0)$ and $\bc(0)$ where the corresponding Gram matrices remain close to the positive definite matrix $\bG(0)$.

\begin{lemma}
\label{lemma_local}
Under Assumptions \ref{assump_positive} and \ref{assump_bounded}, 
there exist $R_a>0$, $R_c > 0$ and $M_c>0$, such that if $\left\|\ba - \ba(0) \right\|_2 \leq R_a$,  $\left\|\bc - \bc(0) \right\|_2 \leq R_c$ and $\left\| \bc_{q}\right\|_{2} \leq M_c$ for all $q \in [m]$, then 
\begin{equation*}
    \left\|\bG - \bG(0)\right\|_2 \leq \frac{\sigma_{\min}}{8}.
\end{equation*}
\end{lemma}

This result is derived from a sensitivity analysis of $\bG$ with respect to the parameters $\ba$ and $\bc$; its proof provided in \Cref{appendix_proof_lemma_local}. 
To guarantee that the Gram matrix remains positive definite along the trajectory of GD, a sufficient condition is that the sequence of parameters does not leave a small neighborhood of the initialization, as identified in \Cref{lemma_local}.
The following lemma confirms our conjecture  regarding the positive-definiteness of the Gram matrices along the GD trajectory under the ``lazy'' training regime of KANs, showing that the sequences generated by GD stay close to their initialization, a phenomenon also observed in MLPs~\cite{du2018gradient,oymak2020toward}. 
This behavior can be interpreted from another perspective. To represent a function as a sum of $m$ functions (i.e., $f \approx f_1 + \ldots + f_m$), any perturbation of $f$ can be  distributed almost equally among the $m$ functions. 
Consequently, small parameter updates still contribute to the expressiveness of the KAN, whereas KANs with smaller $m$ are subject to significant parameter shifts during training.
Thus, this intuitive explanation suggests that wider KANs exhibit faster convergence due to the ``lazy" training property. 
The proof can be found in \Cref{appendix_proof_lemma_next_step_inball}.

\begin{lemma}
\label{lemma_next_step_inball}  
Under Assumptions \ref{assump_positive} and \ref{assump_bounded}, and  if  
\begin{equation*}
        m \gtrsim \frac{n_d^{7} n^3}{\sigma_{\min}^{4}} \left(n_d^2 + \left(\log \frac{m}{\delta} \right)^2 \right) \cdot \left(\log \frac{N}{\delta}\right),
    \end{equation*}
    $\left\| \bs(\tau)\right\|_2 \leq \left(1- \eta \cdot \frac{\sigma_{\min}}{2} \right)^{\tau} \cdot \left\| \bs(0)\right\|_2$, $\left\|\ba(\tau) - \ba(0) \right\|_2 \leq R_a$,  $\left\|\bc(\tau) - \bc(0) \right\|_2 \leq R_c$, and $\left\| \bc_{q}(\tau)\right\|_{2} \leq M_c$ for all $q \in [m]$ and $0 \leq \tau \leq t$, then 
    \begin{equation*}
        \left\|\ba(t+1) - \ba(0) \right\|_2 \leq R_a,\quad
        \left\|\bc(t+1) - \bc(0) \right\|_2 \leq R_c,
    \end{equation*}
    and $\left\| \bc_{q}(\tau)\right\|_{2} \leq M_c$,
    for all $q \in [m]$, where $R_a$, $R_c$, and $M_c$ are as defined in \Cref{lemma_local}. 
\end{lemma}

The next lemma quantifies the error introduced by the discretized approximation of GD in \Cref{eq_gd} to the continuous gradient flow in \Cref{eq_gradient_flow}. 
It shows that the discretization error is small relative to the decrease in the objective function, provided that the step size is moderate and the hidden dimension $m$ is sufficiently large. The  proof can be found in \Cref{appendix_proof_lemma_bound_tau_res}.

\begin{lemma}
\label{lemma_bound_tau_res}
    Let $\chi_{i}(\tau)$ be the error of the first-order approximation at the $i$-th sample, defined as 
    \begin{align}
            \chi_{i}(\tau)
            &:=  s_{i}(\tau\!+\! 1) - s_{i}(\tau) - \left \langle \frac{\partial s_i(\tau)}{\partial \ba}, \ba(\tau+1)\!-\!\ba(\tau)\right\rangle \nonumber\\
            & \qquad - \left \langle \frac{\partial s_i(\tau)}{\partial \bc}, \bc(\tau+1)-\bc(\tau)\right\rangle,\label{eq_error}
       \end{align}
    and set $\bm{\chi}(\tau) = [\chi_1(\tau),\cdots, \chi_{N}(\tau)]^{\top} \in \mathbb{R}^{N}$.
    Under Assumptions \ref{assump_positive} and \ref{assump_bounded}, and the same condition on $m$ as in \Cref{lemma_next_step_inball},
    if $\left\|\ba(\tau)  - \ba(0) \right\|_2 \leq R_a$,  $\left\|\bc(\tau)  - \bc(0) \right\|_2 \leq R_c$ and $\left\| \bc_{q}(\tau)\right\|_{2} \leq M_c$ with $q \in [m]$, then
    \begin{equation*}
        \left\|\bm{\chi}(\tau) \right\|_2 \leq \frac{\eta \sigma_{\min}}{4} \left\| \bs(\tau)\right\|_2.
    \end{equation*}
\end{lemma}

Combining Lemmas \ref{lemma_init_loss} to~\ref{lemma_bound_tau_res}, we establish  that the training loss optimized by GD decays exponentially to zero. 


\begin{theorem}
\label{theorem_gd}
    Under Assumptions \ref{assump_positive} and \ref{assump_bounded}, and under the condition that 
    \begin{equation}
    \label{eq_m_gd}
        m \gtrsim \frac{n_d^{7} n^3}{\sigma_{\min}^{4}} \left(n_d^2 + \left(\log \frac{m}{\delta} \right)^2 \right) \cdot \left(\log \frac{N}{\delta}\right),
    \end{equation}
    then with probability of at least $1-\delta$ over the initialization,   vanilla GD in \Cref{eq_gd} satisfies
    \begin{equation}
        \mathcal{L}(t) \leq \left(1-\eta \frac{\sigma_{\min}}{2} \right)^{t} \mathcal{L}(0), \label{eqn:Lt}
    \end{equation}
where we require the step size $0 < \eta \lesssim \frac{1}{n_d}$.
\end{theorem}

\textbf{Proof Sketch}. Per \Cref{lemma_init_gram}, at initialization, the Gram matrix $\bG(0)$ is positive definite.  \Cref{lemma_local} prescribes neighborhoods for $\ba$ and $\bc$ such that Gram matrices remain positive definite within these regions. For any given time step $t$, if the objective function decays exponentially and the parameters remain within the defined neighborhoods for $0\leq \tau \leq t$, then \Cref{lemma_next_step_inball} guarantees that the updated parameters $\ba(t+1)$ and $\bc(t+1)$ will also stay within these neighborhoods. By controlling the error term arising from the discretization, we can further show that $\mathcal{L}(t+1) \leq \left(1- \eta \frac{\sigma_{\min}}{2}\right) \mathcal{L}(t)$. Thus, by induction, we obtain \Cref{eqn:Lt} since all conditions are satisfied for every $t$.
The   proof is provided in \Cref{appendix_proof_theorem_gd}.

As observed from \Cref{theorem_gd}, GD exhibits {\em linear convergence} when optimizing two-layer KANs with a sufficiently large hidden dimension given by \Cref{eq_m_gd}. This partially explains the phenomenon observed in some recent works~\cite{wang2024kolmogorov,liu2024kan,ss2024chebyshev,aghaei2024rkan}, where gradient-based algorithms achieve near-zero training loss. 
As noted in \Cref{eq_m_gd}, we have higher requirements for the hidden dimension $m$ if more basis functions are considered (i.e., larger $n_d$), which is consistent with the empirical observation from \cite{liu2024kan} that more basis functions may introduce additional optimization challenges.
We do not claim that the dependence of the hidden dimension $m$ on various parameters is  tight but note that  $m$ scales polynomially and logarithmically with the number of basis functions $n_d$, the problem dimension $n$, the minimal positive eigenvalue $\sigma_{\min}$ of the expected Gram matrix, the number of training samples $N$, and the probability of violation $\delta$.  Improving the dependencies of $m$ on the parameters, or showing sharpness, is left for future work.

However, there are still some aspects of KANs that are not fully captured by this analysis.
For example, our results do not emphasize the inherent advantages of KANs over MLPs. 
Although our refined bound on $m$ is competitive vis-\`a-vis existing results for MLPs~\cite{du2018gradient,oymak2020toward}, this improvement mainly results from the adoption of different proof techniques (e.g., refined concentration inequalities, sharper error bounds, and proof by induction in Lemmas \ref{lemma_init_loss}, \ref{lemma_next_step_inball}, and \ref{lemma_bound_tau_res}) rather than structural differences between KANs and MLPs. 
This suggests that the role of basis functions in KANs has  not been fully explored in the current analysis.
We believe that providing a theoretical explanation for the advantages of KANs is an important and promising research direction, providing a deeper understanding of KANs.

\section{Convergence Analysis for SGD}
\label{sec_sgd}
In this section, we analyze convergence of SGD when it is used to optimize the loss function in \Cref{eq_loss}. Compared to \Cref{sec_gd}, where we examined the convergence of GD, this section focuses on a more practical setting using SGD. The SGD algorithm updates trainable parameters according to the following rule
\begin{equation}
\label{eq_sgd}
    \begin{split}
         \ba(t+1) & = \ba(t) -\eta \cdot \frac{\partial \Tilde{\mathcal{L}}}{\partial \ba}(\ba(t),\bc(t)),\\\bc(t+1) & = \bc(t) -\eta \cdot \frac{\partial \Tilde{\mathcal{L}}}{\partial \bc}(\ba(t),\bc(t)),
    \end{split}
\end{equation}
where $\eta > 0$ is the step size, and
\begin{equation}
    \Tilde{\mathcal{L}}(\ba,\bc) = \frac{1}{b} \sum_{i \in \mathcal{I}} \left( f_{i}(\ba,\bc) - y_i\right)^2,
\end{equation}
where $\mathcal{I}$ is the set of indices (mini-batch) sampled uniformly from $[N]$, and $b = |\mathcal{I}|$ denotes the batch size. 
The idea of proof is similar to that in \Cref{sec_gd}, however, the stochasticity introduced by SGD adds additional noise, making it more challenging to control the trajectory of the optimization process.

Here, we highlight some key differences of the analysis of  GD and SGD. 
First, unlike the {\em deterministic} linear convergence of GD established in \Cref{theorem_gd}, we show that SGD achieves linear convergence {\em in expectation} due to its  stochasticity.
Second, the ``lazy'' training behavior of the parameters described in \Cref{lemma_next_step_inball} is not guaranteed for SGD. Instead, we demonstrate that the parameters remain within a neighborhood of their initialization with high probability. 
Third, the theoretical requirements for the hidden dimension and step size are more stringent for SGD to control the stability of iterates in the presence of stochasticity.
The detailed analysis is conducted as follows, and all proofs can be found in \Cref{appendix_proof_sgd}.

\begin{lemma}
\label{lemma_local_sgd}
    Under Assumptions \ref{assump_positive} and \ref{assump_bounded}, there exist $R~\text{and}~\widetilde{M}_c$ such that if $\left\|\ba - \ba(0) \right\|_2 \leq R$, $\left\|\bc - \bc(0) \right\|_2 \leq R$, and $\left\|\bc_q \right\|_2 \leq \widetilde{M}_c$ for all $q \in [m]$, then 
    \begin{equation*}
        \left\|\bG(\ba,\bc) - \bG(\ba(0),\bc(0)) \right\|_2 \leq \frac{\sigma_{\min}}{8}.
    \end{equation*}
\end{lemma}


Let $T$ be the random stopping time of the iterates, such that for $0 \leq t < T$, we have $\left\|\ba(t) - \ba(0) \right\|_2 \leq R/2$, $\left\|\bc(t) - \bc(0) \right\|_2 \leq R/2$, and $\left\|\bc_q(t) \right\|_2 \leq \widetilde{M}_c/2$ for all $q \in [m]$, with at least one of these conditions failing at $t = T$. We denote $\mathcal{F}_{t}$ as the $\sigma$-algebra generated by the sequence of random variables (e.g., stochasticity of mini-batch) up to time step $t$. 
Due to the stochasticity of SGD, we incorporate the indicator function $\bm{1}_{T > t}$ into the conditional expectation to control the trajectory of SGD within the defined neighborhood.

\begin{lemma}
\label{lemma_expect_move}
    Under Assumptions \ref{assump_positive} and \ref{assump_bounded}, for $0 \leq t < T$, we have
    \begin{equation*}
        \mathbb{E}\left[\left\| \ba(t+1) - \ba(t) \right\|_2^2 \cdot \bm{1}_{T > t}|\mathcal{F}_{t}\right] \leq \frac{N}{b} \eta^2 n_d \cdot \mathcal{L}(t),
    \end{equation*}
    and
    \begin{equation*}
         \mathbb{E}\left[\left\| \bc(t+1) - \bc(t)\right\|_2^2 \cdot \bm{1}_{T > t}|\mathcal{F}_{t}\right] \leq \frac{N}{b} \eta^2 n_d \cdot \mathcal{L}(t).
    \end{equation*}
\end{lemma}

The above lemma, whose   proof is provided in \Cref{appendix_proof_lemma_expect_move}, implies that the expected length of step update is proportional to $\frac{N}{b}$.
This result indicates that using a smaller mini-batch size $b$ introduces larger noise, which is consistent with our intuition.
Although $T$ is a stopping time, we can still control $\ba(T)$ and $\bc(T)$ to remain within the neighborhoods in \Cref{lemma_local_sgd} by using a sufficiently small step size, as shown in \Cref{lemma_inball}.
This is one of the key tricks and insights in our proof, where $T$ denotes the stopping time when the parameters reach half their radii of their neighborhoods, while the iterates at time $T$ still stay inside the neighborhood if the step size is small enough, i.e., $\eta \lesssim \frac{b}{N n_d}$. 
This requirement on the step size also agrees with practical observations~\cite{sutskever2013importance,goyal2017accurate,arora2022understanding,yun2022minibatch,Lin2020Dont}, as smaller batch sizes typically prefer smaller step sizes to counteract the increased noise. The   proof is available in \Cref{appendix_proof_lemma_inball}.

\begin{lemma}
\label{lemma_inball}
    Under Assumptions \ref{assump_positive} and \ref{assump_bounded}, if $\eta \lesssim \frac{b}{N n_d}$, then we have
    \begin{equation*}
        \left\|\ba(T) -\ba(0)\right\|_2 \leq R_a,\quad \left\|\bc(T) -\bc(0)\right\|_2 \leq R_c,
    \end{equation*}
    and $\left\|\bc_{q}(T)\right\|_2 \leq \widetilde{M}_c$,
    for all $q \in [m]$. 
\end{lemma}

Next, we refine \Cref{lemma_bound_tau_res} for GD to \Cref{lemma_error_sgd} for SGD, incorporating the stopping time. This refinement is crucial, as we are  concerned with guaranteeing that the parameters remain within a   neighborhood of their initialization, preserving the ``lazy training" property. 
The proof is in \Cref{appendix_proof_lemma_error_sgd}.

\begin{lemma}
\label{lemma_error_sgd}
    Let $\chi_{i}(\tau)$ be the error of the first-order approximation of the $i$-th sample and let 
    $\bm{\chi}(\tau) = [\chi_1(\tau),\cdots, \chi_{N}(\tau)]^{\top} \in \mathbb{R}^{N}$.
    Under Assumptions \ref{assump_positive} and~\ref{assump_bounded}, for $0 \leq t < T$, we have 
    \begin{equation*}
        \mathbb{E}\left[\left\| \bm{\chi}(t)\right\|_2 \cdot \left\| \bs(t)\right\|_2 \cdot \bm{1}_{T > t}|\mathcal{F}_{t}\right] \leq \frac{\eta \sigma_{\min}}{8} \cdot \mathcal{L}(t),
    \end{equation*}
    and
    \begin{equation*}
        \mathbb{E}\left[\left\|\bs(t+1) - \bs(t) \right\|_2^2 \cdot \bm{1}_{T > t}|\mathcal{F}_{t}\right] \leq  \frac{\eta \sigma_{\min}}{8} \cdot \mathcal{L}(t),
    \end{equation*}
    where we require $\eta \lesssim \frac{b}{N} \cdot \frac{\sigma_{\min}}{n_d^{2}}$. 
\end{lemma}

In contrast to the {\em deterministic} decrease of the objective function optimized by GD, the objective function with SGD exhibits sufficient decrease {\em  in expectation}. The  proof of the following lemma is in \Cref{appendix_proof_lemma_convg_sgd}.

\begin{lemma}
    \label{lemma_convg_sgd}
    Under Assumptions \ref{assump_positive} and \ref{assump_bounded}, one-step SGD achieves a   decrease in the objective function in expectation as follows
    \begin{equation*}
    \mathbb{E}\left[\left.\mathcal{L}(t+1)\cdot \bm{1}_{T > t}\right|\mathcal{F}_{t}\right] \leq  \left(1 - \eta \sigma_{\min}\right) \cdot \mathcal{L}(t),
\end{equation*}
and
$   
    \mathbb{E}\left[\mathcal{L}(t+1)\cdot \bm{1}_{T > t}\right] \leq  \left(1 - \eta \sigma_{\min}\right)^{t+1} \cdot \mathcal{L}(0),
$
where $\eta \lesssim \frac{b}{N} \cdot \frac{\sigma_{\min}}{n_d^{2}}$.
\end{lemma}

Recall that the trajectory of GD is guaranteed to stay close to the initialization and achieves linear convergence.
In the stochastic case, however, the event  $\mathcal{E} = \{T = \infty\}$ is not guaranteed to hold due to the noise introduced by SGD. 
Fortunately, in the following lemma, we prove that  $\mathcal{E}$ holds with high probability, if the hidden dimension $m$ is sufficiently large. The proof is provided in \Cref{appendix_proof_lemma_event_never_leave_ball}.

\begin{lemma}
\label{lemma_event_never_leave_ball}
    Let $\mathcal{E}$ be the event $\{T=\infty\}$. Under Assumptions \ref{assump_positive} and \ref{assump_bounded}, and the condition that
    \begin{equation*}
    \begin{split}
        m &\gtrsim  \frac{N^3  n_d^{11} n^3}{b^3 \sigma_{\min}^{8} \Tilde{\delta}^{6}} \cdot \left(\log \frac{N}{\delta}\right)^{3}\\
        &\qquad + \frac{N n_d^{7} n^3}{b \sigma_{\min}^4 \Tilde{\delta}^6} \cdot \left(\log \frac{m}{\delta}\right)^{2} \cdot \left(\log \frac{N}{\delta}\right),
    \end{split}
\end{equation*}
    the event $\mathcal{E}$ holds with probability at least $1-\Tilde{\delta}$.
\end{lemma}

Summarizing Lemmas \ref{lemma_local_sgd} to~\ref{lemma_event_never_leave_ball}, we deduce that SGD converges linearly in expectation, provided that the hidden dimension $m$ is sufficiently large.

\begin{theorem}
\label{theorem_sgd}   
Under Assumptions \ref{assump_positive} and \ref{assump_bounded}, and if
\begin{equation}
\label{eq_m_sgd}
    \begin{split}
        m &\gtrsim  \frac{N^3  n_d^{11} n^3}{b^3 \sigma_{\min}^{8} \Tilde{\delta}^{6}} \cdot \left(\log \frac{N}{\delta}\right)^{3}\\
        & \qquad+ \frac{N n_d^{7} n^3}{b \sigma_{\min}^4 \Tilde{\delta}^6} \cdot \left(\log \frac{m}{\delta}\right)^{2} \cdot \left(\log \frac{N}{\delta}\right),
    \end{split}
\end{equation}
then with probability   at least $1-\delta$ over the initialization, and there exists an event $\mathcal{E}=\{T=\infty\}$ which holds with probability of $1-\Tilde{\delta}$, SGD   satisfies
\begin{equation*}
    \mathbb{E}\left[\mathcal{L}(t) \cdot \bm{1}_{\mathcal{E}}\right] \leq  \left(1 - \eta \sigma_{\min}\right)^{t} \cdot \mathcal{L}(0),
\end{equation*}
where we require $\eta \lesssim \frac{b}{N} \cdot \frac{\sigma_{\min}}{n_d^{2}}$.
\end{theorem}


Comparing the condition in \Cref{eq_m_sgd} with \Cref{eq_m_gd}, the second part of the right-hand side of \Cref{eq_m_sgd} for SGD matches the result of \Cref{eq_m_gd} for GD. 
However, the additional term in \Cref{eq_m_sgd} arises due to the stochasticity of SGD.
These results demonstrate why training KANs with SGD can successfully achieve near-zero training loss, making the analysis more practical and informative compared to GD.
However, as discussed following \Cref{theorem_gd}, the potential limitations of the results for GD also exist for SGD. For example, the number of basis functions $n_d$ does not contribute in the derived convergence results, and the structural superiority of KANs over MLPs is not fully captured. 
Despite these limitations, this is the first work to theoretically establish the convergence of SGD for training KANs, providing a theoretical explanation for the effectiveness of SGD in KAN implementations and offering confidence in the use of SGD for optimizing KANs in practice.

\section{Physics-Informed KANs}
\label{sec_pikan}
Physics-informed machine learning is one of the most promising and potential applications for KANs~\cite{wang2024kolmogorov,aghaei2024kantrol,toscano2024inferring,rigas2024adaptive}.  In this section, we focus on the convergence analysis of GD and SGD for physics-informed KANs. 
Consider the following second-order partial differential equation (PDE):
\begin{align*}
         \mathcal{D}[u(\bx)]& =  v(\bx), \quad \bx \in \Omega,\\
         u(\bx)& = \Bar{v}(\bx), \quad \bx \in \partial \Omega,
\end{align*}
where $\Omega \subseteq \mathbb{R}^{n}$ is the domain of interest, $\partial \Omega$ is its boundary, and the first component $x_1$ of $\bx$ denotes the temporal domain, while the remaining components represent the spatial variables.
We denote the differential operator as $\mathcal{D}[u(\bx)]$, where $\mathcal{D}[u(\bx)] := \frac{\partial u(\bx)}{\partial x_{1}} - \sum_{i,j=2}^{n} h_{i,j}(\bx) \cdot \frac{\partial^2 u(\bx)}{\partial x_{i} \partial x_{j}} - \sum_{i=2}^{n} g_{i}(\bx) \cdot \frac{\partial u(\bx)}{\partial x_{i}} - l(\bx) \cdot u(\bx)$.
Given training samples $\{(\bx_{i}, v_i)\}_{i=1}^{N_1}$ in the interior domain and $\{(\Bar{\bx}_{i}, u_i)\}_{i=1}^{N_2}$ on the boundary, the loss function for physics-informed KANs is formulated as
\begin{equation}
\label{eq_loss_kan_pde}
        \mathcal{L}^{\text{PDE}}(\ba,\bc)
        =  \frac{1}{N_1}\sum_{i=1}^{N_1} \left( \mathcal{D}[f(\bx_{i};\ba,\bc)] - v_i \right)^2
         + \frac{1}{N_2}\sum_{i=1}^{N_2} \left( f(\Bar{\bx}_{i};\ba,\bc) - u_i \right)^2.
\end{equation}
In SGD, the mini-batch loss is given by
\begin{equation}
\label{eq_loss_kan_pde_sgd}
        \Tilde{\mathcal{L}}^{\text{PDE}}(\ba,\bc) =   \frac{1}{b_1}\sum_{i \in \mathcal{I}} \left( \mathcal{D}[f(\bx_{i};\ba,\bc)] - v_i \right)^2
        + \frac{1}{b_2}\sum_{i \in \Bar{\mathcal{I}} } \left( f(\Bar{\bx}_{i};\ba,\bc) - u_i \right)^2,
\end{equation}
where $\mathcal{I}$ and $\Bar{\mathcal{I}}$ are the sets of indices for mini-batches sampled from the interior and boundary, with batch sizes $b_1$ and $b_2$, respectively. We assume that $\frac{b_1}{N_1} = \frac{b_2}{N_2}$.
Due to   space limits, we omit further details here and provide all technical derivations in \Cref{appendix_proof_pikan}.

\begin{theorem}
\label{theorem_gd_pde}
    Under Assumptions \ref{assump_positive} and \ref{assump_bounded}, if
    \begin{equation}
    \label{eq_m_gd_pde}
        m \gtrsim \frac{n_d^{10} n^{15}}{\sigma_{\min}^{4}} \left(n_d^{6} n^{4} + \left(\log \frac{m}{\delta} \right)^6 \right) \cdot \left(\log \frac{N_1 + N_2}{\delta}\right),
    \end{equation}
    then with probability of at least $1-\delta$ over the initialization, the vanilla GD for physics-informed KANs (\Cref{eq_loss_kan_pde}) satisfies
    \begin{equation*}
        \mathcal{L}^{\text{PDE}}(t) \leq \left(1-\eta \frac{\sigma_{\min}}{2} \right)^{t} \mathcal{L}^{\text{PDE}}(0),
    \end{equation*}
where we require $0 < \eta \lesssim \frac{1}{n_d^3 n^4}$.
\end{theorem}

\begin{theorem}
\label{theorem_sgd_pde}   
Under Assumptions \ref{assump_positive} and \ref{assump_bounded}, if
\begin{equation}
\label{eq_m_sgd_pde}
    \begin{split}
        m&\gtrsim  \frac{N_1^7  n_d^{40} n^{63}}{b_1^7 \sigma_{\min}^{16} \Tilde{\delta}^{14}} \cdot \left(\log \frac{N_1 + N_2}{\delta}\right)^{7}\\
        & + \frac{N_1 n_d^{10} n^{15}}{b_1 \sigma_{\min}^4 \Tilde{\delta}^{14}} \cdot \left(n_d^6 n^4\! + \!\left(\log \frac{m}{\delta}\right)^{6} \right) \cdot \left(\log \frac{N_1 \!+\! N_2}{\delta}\right),
    \end{split}
\end{equation}
then with probability of at least $1-\delta$ over the initialization, and there exists an event $\mathcal{E}$ which holds with probability of $1-\Tilde{\delta}$ such that SGD for physics-informed KANs (Equations \ref{eq_loss_kan_pde} and \ref{eq_loss_kan_pde_sgd}) satisfies
\begin{equation*}
    \mathbb{E}\left[\mathcal{L}^{\text{PDE}}(t) \cdot \bm{1}_{\mathcal{E}}\right] \leq  \left(1 - \eta \sigma_{\min}\right)^{t} \cdot \mathcal{L}^{\text{PDE}}(0),
\end{equation*}
where we require $\eta \lesssim \frac{b_1 \sigma_{\min}}{N_1 n_d^6 n^8}$.
\end{theorem}

Comparing Theorems \ref{theorem_gd_pde} and \ref{theorem_sgd_pde} to Theorems \ref{theorem_gd} and \ref{theorem_sgd}, we have higher requirements on the hidden dimension of physics-informed KANs, due to the additional challenges introduced by the differential operator $\mathcal{D}$. However, the current results do not reflect the superior performance of KANs in physics-informed tasks compared to MLPs.
Moreover, while our results are valid for linear (and even higher-order) PDEs, they may not extend to certain nonlinear PDEs. In the cases of nonlinear PDEs, the Gram matrix $\bG(0)$ does not approximate $\bG^{\infty}$ and instead converges to a random variable rather than a deterministic matrix. This behavior differs from \Cref{lemma_init_gram}, which holds for linear PDEs, where $\bG(t)$ consistently stays close to $\bG^{\infty}$. Although \cite{bonfanti2024challenges} suggests techniques such as Newton’s method to address this issue, Newton’s method is not widely adopted for training KANs due to its high computational complexity. 
We believe that providing a theoretical explanation and deeper analysis of physics-informed KANs, especially for broader classes of PDEs, is essential for understanding their advantages and behavior in these applications.

\section{Conclusion and Future Works}
\label{sec_conclusion}
In this paper,  using the neural tangent kernel perspective, we conduct  rigorous convergence analyses for GD and SGD in optimizing KANs for regression and physics-informed tasks. 
We prove that GD achieves global linear convergence, while SGD enjoys global linear convergence in expectation, provided that the hidden dimension is sufficiently large. 
Our work is the first to theoretically explain the empirical observation that SGD   achieves near-zero training loss when training KANs in various tasks, contributing to a deeper understanding of the behavior and effectiveness of KANs. 
We also discuss some limitations of our current results.

There are several promising directions for future investigation of KANs. First and foremost is the exploration of the inherent and intricate structural advantages of KANs over MLPs, especially in various scientific applications (e.g., physics-informed KANs). While we derived the convergence analysis of SGD for KANs, our results do not fully capture the structural superiority of KANs, making it a compelling and challenging topic for further and deeper research.

Second, our results involve assumptions on the smoothness and boundedness of the basis and transformation functions (\Cref{assump_bounded}). Although these assumptions are realizable for many variants of KANs, they do not fully capture the influence of different basis functions on the model's behaviours. Despite the significant variation in performances observed with different basis functions, our current analyses do not reflect these differences. Therefore, establishing theoretical guidelines for selecting appropriate basis functions is an essential and open area of research.

Lastly, although we have proven that SGD is effective in training KANs for regression and physics-informed tasks, exploring other optimization algorithms specifically designed to leverage the unique structure of KANs could yield further performance improvements. As a brand new model, KANs have shown great potential across various applications but still require more in-depth exploration. This work is only the beginning, and there is much more to discover and understand about KANs.

\newpage 
\bibliographystyle{plain}
\bibliography{main}

\begin{thebibliography}{10}

\bibitem{abueidda2024deepokan}
Diab~W Abueidda, Panos Pantidis, and Mostafa~E Mobasher.
\newblock Deep{OKAN}: Deep operator network based on {K}olmogorov {A}rnold networks for mechanics problems.
\newblock {\em arXiv preprint arXiv:2405.19143}, 2024.

\bibitem{aghaei2024fkan}
Alireza~Afzal Aghaei.
\newblock fkan: Fractional {K}olmogorov--{A}rnold {N}etworks with trainable {J}acobi basis functions.
\newblock {\em arXiv preprint arXiv:2406.07456}, 2024.

\bibitem{aghaei2024kantrol}
Alireza~Afzal Aghaei.
\newblock {KAN}trol: A physics-informed {K}olmogorov--{A}rnold network framework for solving multi-dimensional and fractional optimal control problems.
\newblock {\em arXiv preprint arXiv:2409.06649}, 2024.

\bibitem{aghaei2024rkan}
Alireza~Afzal Aghaei.
\newblock {rKAN}: Rational {K}olmogorov--{A}rnold networks.
\newblock {\em arXiv preprint arXiv:2406.14495}, 2024.

\bibitem{allen2019convergence}
Zeyuan Allen-Zhu, Yuanzhi Li, and Zhao Song.
\newblock A convergence theory for deep learning via over-parameterization.
\newblock In {\em International conference on machine learning}, pages 242--252. PMLR, 2019.

\bibitem{arora2022understanding}
Sanjeev Arora, Zhiyuan Li, and Abhishek Panigrahi.
\newblock Understanding gradient descent on the edge of stability in deep learning.
\newblock In {\em International Conference on Machine Learning}, pages 948--1024. PMLR, 2022.

\bibitem{bodner2024convolutional}
Alexander~Dylan Bodner, Antonio~Santiago Tepsich, Jack~Natan Spolski, and Santiago Pourteau.
\newblock Convolutional kolmogorov-arnold networks.
\newblock {\em arXiv preprint arXiv:2406.13155}, 2024.

\bibitem{bonfanti2024challenges}
Andrea Bonfanti, Giuseppe Bruno, and Cristina Cipriani.
\newblock The challenges of the nonlinear regime for physics-informed neural networks.
\newblock {\em arXiv preprint arXiv:2402.03864}, 2024.

\bibitem{bozorgasl2024wav}
Zavareh Bozorgasl and Hao Chen.
\newblock Wav-{KAN}: Wavelet {K}olmogorov--{A}rnold {N}etworks.
\newblock {\em arXiv preprint arXiv:2405.12832}, 2024.

\bibitem{braun2009constructive}
J{\"u}rgen Braun and Michael Griebel.
\newblock On a constructive proof of kolmogorov’s superposition theorem.
\newblock {\em Constructive Approximation}, 30:653--675, 2009.

\bibitem{carmon2018accelerated}
Yair Carmon, John~C Duchi, Oliver Hinder, and Aaron Sidford.
\newblock Accelerated methods for nonconvex optimization.
\newblock {\em SIAM Journal on Optimization}, 28(2):1751--1772, 2018.

\bibitem{chen2024gaussian}
Andrew~Siyuan Chen.
\newblock Gaussian process {K}olmogorov--{A}rnold networks.
\newblock {\em arXiv preprint arXiv:2407.18397}, 2024.

\bibitem{cheon2024demonstrating}
Minjong Cheon.
\newblock Demonstrating the efficacy of {K}olmogorov-{A}rnold networks in vision tasks.
\newblock {\em arXiv preprint arXiv:2406.14916}, 2024.

\bibitem{chizat2018global}
Lenaic Chizat and Francis Bach.
\newblock On the global convergence of gradient descent for over-parameterized models using optimal transport.
\newblock {\em Advances in Neural Information Processing Systems}, 31, 2018.

\bibitem{drokin2024kolmogorov}
Ivan Drokin.
\newblock Kolmogorov--{A}rnold convolutions: Design principles and empirical studies.
\newblock {\em arXiv preprint arXiv:2407.01092}, 2024.

\bibitem{du2018gradient}
Simon~S. Du, Xiyu Zhai, Barnabas Poczos, and Aarti Singh.
\newblock Gradient descent provably optimizes over-parameterized neural networks.
\newblock In {\em International Conference on Learning Representations}, 2019.

\bibitem{duchi2011adaptive}
John Duchi, Elad Hazan, and Yoram Singer.
\newblock Adaptive subgradient methods for online learning and stochastic optimization.
\newblock {\em Journal of Machine Learning Research}, 12(7), 2011.

\bibitem{gao2023gradient}
Yihang Gao, Yiqi Gu, and Michael Ng.
\newblock Gradient descent finds the global optima of two-layer physics-informed neural networks.
\newblock In {\em International Conference on Machine Learning}, pages 10676--10707. PMLR, 2023.

\bibitem{genet2024temporal}
Remi Genet and Hugo Inzirillo.
\newblock A temporal {K}olmogorov--{A}rnold transformer for time series forecasting.
\newblock {\em arXiv preprint arXiv:2406.02486}, 2024.

\bibitem{genet2024tkan}
Remi Genet and Hugo Inzirillo.
\newblock {TKAN}: {T}emporal {K}olmogorov--{A}rnold networks.
\newblock {\em arXiv preprint arXiv:2405.07344}, 2024.

\bibitem{goyal2017accurate}
Priya Goyal, Piotr Doll{\'a}r, Ross Girshick, Pieter Noordhuis, Lukasz Wesolowski, Aapo Kyrola, Andrew Tulloch, Yangqing Jia, and Kaiming He.
\newblock Accurate, large minibatch {SGD}: training {ImageNet} in 1 hour.
\newblock {\em arXiv preprint arXiv:1706.02677}, 2017.

\bibitem{guo2024kan}
Haihong Guo, Fengxin Li, Jiao Li, and Hongyan Liu.
\newblock {KAN} vs {MLP} for offline reinforcement learning.
\newblock {\em arXiv preprint arXiv:2409.09653}, 2024.

\bibitem{han2024kan4tsf}
Xiao Han, Xinfeng Zhang, Yiling Wu, Zhenduo Zhang, and Zhe Wu.
\newblock {KAN4TSF}: Are {KAN} and {KAN}-based models effective for time series forecasting?
\newblock {\em arXiv preprint arXiv:2408.11306}, 2024.

\bibitem{he2015delving}
Kaiming He, Xiangyu Zhang, Shaoqing Ren, and Jian Sun.
\newblock Delving deep into rectifiers: Surpassing human-level performance on {I}magenet classification.
\newblock In {\em Proceedings of the IEEE International Conference on Computer Vision}, pages 1026--1034, 2015.

\bibitem{hou2024comprehensive}
Yuntian Hou, Jinheng Wu, Xiaohang Feng, et~al.
\newblock A comprehensive survey on {K}olmogorov {A}rnold networks (kan).
\newblock {\em arXiv preprint arXiv:2407.11075}, 2024.

\bibitem{howard2024finite}
Amanda~A Howard, Bruno Jacob, Sarah~H Murphy, Alexander Heinlein, and Panos Stinis.
\newblock Finite basis {K}olmogorov--{A}rnold networks: domain decomposition for data-driven and physics-informed problems.
\newblock {\em arXiv preprint arXiv:2406.19662}, 2024.

\bibitem{jacot2018neural}
Arthur Jacot, Franck Gabriel, and Cl{\'e}ment Hongler.
\newblock Neural tangent kernel: Convergence and generalization in neural networks.
\newblock {\em Advances in Neural Information Processing Systems}, 31, 2018.

\bibitem{kich2024kolmogorov}
Victor~Augusto Kich, Jair~Augusto Bottega, Raul Steinmetz, Ricardo~Bedin Grando, Ayano Yorozu, and Akihisa Ohya.
\newblock {K}olmogorov--{A}rnold network for online reinforcement learning.
\newblock {\em arXiv preprint arXiv:2408.04841}, 2024.

\bibitem{kingma2015adam}
Diederik~P Kingma and Jimmy Ba.
\newblock Adam: A method for stochastic optimization.
\newblock {\em International Conference for Learning Representations (ICLR)}, 2015.

\bibitem{klambauer2017self}
G{\"u}nter Klambauer, Thomas Unterthiner, Andreas Mayr, and Sepp Hochreiter.
\newblock Self-normalizing neural networks.
\newblock {\em Advances in Neural Information Processing Systems}, 30, 2017.

\bibitem{knottenbelt2024coxkan}
William Knottenbelt, Zeyu Gao, Rebecca Wray, Woody~Zhidong Zhang, Jiashuai Liu, and Mireia Crispin-Ortuzar.
\newblock {CoxKAN}: {K}olmogorov--{A}rnold networks for interpretable, high-performance survival analysis.
\newblock {\em arXiv preprint arXiv:2409.04290}, 2024.

\bibitem{kolmogorov1957representation}
Andrei~Nikolaevich Kolmogorov.
\newblock On the representation of continuous functions of many variables by superposition of continuous functions of one variable and addition.
\newblock In {\em Doklady Akademii Nauk}, volume 114, pages 953--956. Russian Academy of Sciences, 1957.

\bibitem{kolmogorov1961representation}
Andre{\u\i}~Nikolaevich Kolmogorov.
\newblock {\em On the representation of continuous functions of several variables by superpositions of continuous functions of a smaller number of variables}.
\newblock American Mathematical Society, 1961.

\bibitem{kuchibhotla2022moving}
Arun~Kumar Kuchibhotla and Abhishek Chakrabortty.
\newblock Moving beyond sub-gaussianity in high-dimensional statistics: Applications in covariance estimation and linear regression.
\newblock {\em Information and Inference: A Journal of the IMA}, 11(4):1389--1456, 2022.

\bibitem{li2024u}
Chenxin Li, Xinyu Liu, Wuyang Li, Cheng Wang, Hengyu Liu, and Yixuan Yuan.
\newblock {U-KAN} makes strong backbone for medical image segmentation and generation.
\newblock {\em arXiv preprint arXiv:2406.02918}, 2024.

\bibitem{li2019convergence}
Xiaoyu Li and Francesco Orabona.
\newblock On the convergence of stochastic gradient descent with adaptive stepsizes.
\newblock In {\em The 22nd International Conference on Artificial Intelligence and Statistics}, pages 983--992. PMLR, 2019.

\bibitem{Lin2020Dont}
Tao Lin, Sebastian~U. Stich, Kumar~Kshitij Patel, and Martin Jaggi.
\newblock Don't use large mini-batches, use local {SGD}.
\newblock In {\em International Conference on Learning Representations}, 2020.

\bibitem{liu2024kan2}
Ziming Liu, Pingchuan Ma, Yixuan Wang, Wojciech Matusik, and Max Tegmark.
\newblock Kan 2.0: Kolmogorov-arnold networks meet science.
\newblock {\em arXiv preprint arXiv:2408.10205}, 2024.

\bibitem{liu2024kan}
Ziming Liu, Yixuan Wang, Sachin Vaidya, Fabian Ruehle, James Halverson, Marin Solja{\v{c}}i{\'c}, Thomas~Y Hou, and Max Tegmark.
\newblock {KAN}: Kolmogorov--{A}rnold networks.
\newblock {\em arXiv preprint arXiv:2404.19756}, 2024.

\bibitem{nagai2024kolmogorov}
Yuki Nagai and Masahiko Okumura.
\newblock {K}olmogorov--{A}rnold networks in molecular dynamics.
\newblock {\em arXiv preprint arXiv:2407.17774}, 2024.

\bibitem{nesterov2006cubic}
Yurii Nesterov and Boris~T Polyak.
\newblock Cubic regularization of {N}ewton method and its global performance.
\newblock {\em Mathematical Programming}, 108(1):177--205, 2006.

\bibitem{oymak2020toward}
Samet Oymak and Mahdi Soltanolkotabi.
\newblock Toward moderate overparameterization: Global convergence guarantees for training shallow neural networks.
\newblock {\em IEEE Journal on Selected Areas in Information Theory}, 1(1):84--105, 2020.

\bibitem{polar2021deep}
Andrew Polar and Michael Poluektov.
\newblock A deep machine learning algorithm for construction of the {K}olmogorov--{A}rnold representation.
\newblock {\em Engineering Applications of Artificial Intelligence}, 99:104137, 2021.

\bibitem{qiu2024relu}
Qi~Qiu, Tao Zhu, Helin Gong, Liming Chen, and Huansheng Ning.
\newblock {ReLU-KAN}: New {K}olmogorov--{A}rnold networks that only need matrix addition, dot multiplication, and {ReLU}.
\newblock {\em arXiv preprint arXiv:2406.02075}, 2024.

\bibitem{reddi2019convergence}
Sashank~J Reddi, Satyen Kale, and Sanjiv Kumar.
\newblock On the convergence of {A}dam and beyond.
\newblock {\em International Conference for Learning Representations}, 2019.

\bibitem{rigas2024adaptive}
Spyros Rigas, Michalis Papachristou, Theofilos Papadopoulos, Fotios Anagnostopoulos, and Georgios Alexandridis.
\newblock Adaptive training of grid-dependent physics-informed {K}olmogorov--{A}rnold networks.
\newblock {\em arXiv preprint arXiv:2407.17611}, 2024.

\bibitem{seydi2024exploring}
Seyd~Teymoor Seydi.
\newblock Exploring the potential of polynomial basis functions in {K}olmogorov--{A}rnold {N}etworks: A comparative study of different groups of polynomials.
\newblock {\em arXiv preprint arXiv:2406.02583}, 2024.

\bibitem{soltanolkotabi2017learning}
Mahdi Soltanolkotabi.
\newblock Learning {ReLU}s via gradient descent.
\newblock {\em Advances in Neural Information Processing Systems}, 30, 2017.

\bibitem{soltanolkotabi2018theoretical}
Mahdi Soltanolkotabi, Adel Javanmard, and Jason~D Lee.
\newblock Theoretical insights into the optimization landscape of over-parameterized shallow neural networks.
\newblock {\em IEEE Transactions on Information Theory}, 65(2):742--769, 2018.

\bibitem{soudry2016no}
Daniel Soudry and Yair Carmon.
\newblock No bad local minima: Data independent training error guarantees for multilayer neural networks.
\newblock {\em arXiv preprint arXiv:1605.08361}, 2016.

\bibitem{ss2024chebyshev}
Sidharth SS.
\newblock Chebyshev polynomial-based {K}olmogorov--{A}rnold networks: An efficient architecture for nonlinear function approximation.
\newblock {\em arXiv preprint arXiv:2405.07200}, 2024.

\bibitem{sutskever2013importance}
Ilya Sutskever, James Martens, George Dahl, and Geoffrey Hinton.
\newblock On the importance of initialization and momentum in deep learning.
\newblock In {\em International Conference on Machine Learning}, pages 1139--1147. PMLR, 2013.

\bibitem{ta2024fc}
Hoang-Thang Ta, Duy-Quy Thai, Abu Bakar~Siddiqur Rahman, Grigori Sidorov, and Alexander Gelbukh.
\newblock {FC-KAN}: Function combinations in {K}olmogorov--{A}rnold {N}etworks.
\newblock {\em arXiv preprint arXiv:2409.01763}, 2024.

\bibitem{toscano2024inferring}
Juan~Diego Toscano, Theo K{\"a}ufer, Martin Maxey, Christian Cierpka, and George~Em Karniadakis.
\newblock Inferring turbulent velocity and temperature fields and their statistics from {L}agrangian velocity measurements using physics-informed {K}olmogorov-{A}rnold networks.
\newblock {\em arXiv preprint arXiv:2407.15727}, 2024.

\bibitem{vaswani2019fast}
Sharan Vaswani, Francis Bach, and Mark Schmidt.
\newblock Fast and faster convergence of {SGD} for over-parameterized models and an accelerated perceptron.
\newblock In {\em The 22nd international conference on artificial intelligence and statistics}, pages 1195--1204. PMLR, 2019.

\bibitem{wang2022and}
Sifan Wang, Xinling Yu, and Paris Perdikaris.
\newblock When and why {PINN}s fail to train: A neural tangent kernel perspective.
\newblock {\em Journal of Computational Physics}, 449:110768, 2022.

\bibitem{wang2024kolmogorov}
Yizheng Wang, Jia Sun, Jinshuai Bai, Cosmin Anitescu, Mohammad~Sadegh Eshaghi, Xiaoying Zhuang, Timon Rabczuk, and Yinghua Liu.
\newblock Kolmogorov {A}rnold {I}nformed neural network: {A} physics-informed deep learning framework for solving pdes based on {K}olmogorov {A}rnold networks.
\newblock {\em arXiv preprint arXiv:2406.11045}, 2024.

\bibitem{xie2017diverse}
Bo~Xie, Yingyu Liang, and Le~Song.
\newblock Diverse neural network learns true target functions.
\newblock In {\em The 20th International Conference on Artificial Intelligence and Statistics}, pages 1216--1224. PMLR, 2017.

\bibitem{xu2024kolmogorov}
Kunpeng Xu, Lifei Chen, and Shengrui Wang.
\newblock {K}olmogorov--{A}rnold networks for time series: Bridging predictive power and interpretability.
\newblock {\em arXiv preprint arXiv:2406.02496}, 2024.

\bibitem{xu2024convergence2}
Xianliang Xu, Ting Du, Wang Kong, Ye~Li, and Zhongyi Huang.
\newblock Convergence analysis of natural gradient descent for over-parameterized physics-informed neural networks.
\newblock {\em arXiv preprint arXiv:2408.00573}, 2024.

\bibitem{xu2024convergence}
Xianliang Xu, Zhongyi Huang, and Ye~Li.
\newblock Convergence of implicit gradient descent for training two-layer physics-informed neural networks.
\newblock {\em arXiv preprint arXiv:2407.02827}, 2024.

\bibitem{yu2024kan}
Runpeng Yu, Weihao Yu, and Xinchao Wang.
\newblock {KAN} or {MLP}: A fairer comparison.
\newblock {\em arXiv preprint arXiv:2407.16674}, 2024.

\bibitem{yun2022minibatch}
Chulhee Yun, Shashank Rajput, and Suvrit Sra.
\newblock Minibatch vs local {SGD} with shuffling: Tight convergence bounds and beyond.
\newblock In {\em International Conference on Learning Representations}, 2022.

\bibitem{zhang2021understanding}
Chiyuan Zhang, Samy Bengio, Moritz Hardt, Benjamin Recht, and Oriol Vinyals.
\newblock Understanding deep learning (still) requires rethinking generalization.
\newblock {\em Communications of the ACM}, 64(3):107--115, 2021.

\bibitem{zhang2024graphkan}
Fan Zhang and Xin Zhang.
\newblock {GraphKAN}: {E}nhancing feature extraction with graph kolmogorov arnold networks.
\newblock {\em arXiv preprint arXiv:2406.13597}, 2024.

\end{thebibliography}

\newpage
\appendix

\section{Proofs for \Cref{sec_gd}}
\label{appendix_proof_gd}
\subsection{Proof for \Cref{lemma_positive}}
\label{appendix_proof_lemma_positive}
We prove this result by contradiction. Suppose that the expected Gram matrix $\bG^{\infty}$ is not strictly definite, then columns of $\bm{D}$ must be linearly dependent for almost all values of $\ba$ and $\bc$ (except for a set of zero-measure). Due to the continuity of $\bm{D}$ with respect to the variables $\ba$ and $\bc$, this almost-everywhere linear dependence would imply that the columns of $\bm{D}$ are linearly dependent everywhere. 
Let the columns of $\bm{D}$ be denoted as $[\bm{d}_1, \bm{d}_2, \cdots, \bm{d}_{N}]$. This means that there exists a nonzero vector $[\alpha_1,\alpha_2, \cdots, \alpha_N]$ such that 
\begin{equation*}
    \alpha_1 \bm{d}_1 + \alpha_2 \bm{d}_2 + \cdots + \alpha_{N} \bm{d}_{N} = 0,
\end{equation*}
for all $\ba$ and $\bc$. Consequently, it implies that $\alpha_1 \frac{\partial f_1}{\partial \bc} + \alpha_2 \frac{\partial f_2}{\partial \bc} + \cdots + \alpha_{N} \frac{\partial f_N}{\partial \bc} = 0,$ which is equivalent to
\begin{equation}
\label{eq_appx_1}
    \begin{split}
        & \alpha_1 \frac{\partial f_1}{\partial c_{q,k}} + \alpha_2 \frac{\partial f_2}{\partial c_{q,k}} + \cdots + \alpha_{N} \frac{\partial f_N}{\partial c_{q,k}} \\
        = &\sum_{i=1}^{N} \alpha_i b_{k}\left(\phi \left( \sum_{p_1=1}^{n}\sum_{k_1=1}^{n_d} a_{p_1,q,k_1} b_{k_1}\left(x_{i,p_1}\right)\right) \right)\\
        = & 0,
    \end{split}
\end{equation}
for all $q \in [m]$ and $k \in [n_d]$. Since $\{b_k\}_{k=1}^{n_d}$ is the set of polynomials of degree less than $n_d$, linear combination of constant function $b_1$ and 1-degree polynomial $b_2$ implies that
\begin{equation*}
    \sum_{i=1}^{N} \alpha_i \phi \left( \sum_{p_1=1}^{n}\sum_{k_1=1}^{n_d} a_{p_1,q,k_1} b_{k_1}\left(x_{i,p_1}\right)\right)=0.
\end{equation*}
Taking derivatives of the left-hand-side of the above equality with respect to $a_{p_1,q,k_1}$ at $\ba=\bm{0}$, we obtain 
\begin{equation*}
    \sum_{i=1}^{N} \alpha_i \phi^{\prime} \left(0\right) b_{k_1}(x_{i,p_1})=0.
\end{equation*}
This equality implies that $\{b_{k_1}(x_{i,p_1})\}_{i=1}^{N}$ are linearly dependent for all $k_1 \in [n_d]$ and $p_1 \in [n]$ under the linear weights $[\alpha_1, \alpha_2, \cdots, \alpha_{N}]$. By taking arbitrary higher-order derivatives (e.g., $d$-th order derivatives), we similarly obtain
\begin{equation*}
    \sum_{i=1}^{N} \alpha_i \phi^{(d)} \left(0\right) b_{k_1}(x_{i,p_1})^{d}=0.
\end{equation*}
The property of the Vandermonde matrix implies the set $\{b_{k_1}(x_{i,p_1})^{d}\}_{i=1}^{N}$ cannot be linearly dependent, given that all training samples $\{\bm{x}_{i}\}_{i=1}^{N}$ are distinct. This contradicts the earlier assumption of linear dependence.

Furthermore, if the transformation function $\phi$ is not applied, then according to Equation (\ref{eq_appx_1}), we have
\begin{equation*}
    \sum_{i=1}^{N} \alpha_i b_{k}\left( \sum_{p_1=1}^{n}\sum_{k_1=1}^{n_d} a_{p_1,q,k_1} b_{k_1}\left(x_{i,p_1}\right)\right) = 0,
\end{equation*}
for all $q \in [m]$ and $k \in [n_d]$. Taking $(k-1)$-th order derivatives with respect to $a_{p_1,q,k_1}$ at $\ba = \bm{0}$, we obtain that 
\begin{equation*}
    \sum_{i=1}^{N} \alpha_i b_{k_1}(x_{i,p_1})^{k-1} = 0,
\end{equation*}
for $k \in [n_d]$, $k_1 \in [n_d]$, and $p_1 \in [n]$. It further implies that 
\begin{equation*}
    \{[1, \bm{x}_{i}^{\top}, \bm{x}_{i}^{2 \top}, \cdots, \bm{x}_{i}^{(n_d-1)^2 \top}]\}_{i=1}^{N}
\end{equation*}
are linearly dependent, which contradicts our requirements for training samples. 
We can employ similar techniques to show by contradiction that $\alpha_1 \frac{\partial f_1}{\partial \bc} + \alpha_2 \frac{\partial f_2}{\partial \bc} + \cdots + \alpha_{N} \frac{\partial f_N}{\partial \bc} \neq 0$ for some $(\ba,\bc)$ of nonzero measure.

Moreover, the $(i,j)$-th element of the Gram matrix $\bG$ can be expressed as
\begin{equation*}
    G_{i,j} = \left\langle \frac{\partial s_{i}}{\partial \ba}, \frac{\partial s_{j}}{\partial \ba}\right \rangle + \left\langle \frac{\partial s_{i}}{\partial \bc}, \frac{\partial s_{j}}{\partial \bc}\right \rangle,
\end{equation*}
where
\begin{equation*}
    \begin{split}
        & \left\langle \frac{\partial s_{i}}{\partial \ba}, \frac{\partial s_{j}}{\partial \ba}\right \rangle \\
        = & \sum_{q=1}^{m} \sum_{p=1}^{n} \sum_{k=1}^{n_d} \frac{\partial s_i}{\partial a_{p,q,k}} \cdot \frac{\partial s_j}{\partial a_{p,q,k}} \\
        = & \frac{1}{mN} \sum_{q=1}^{m} \sum_{p=1}^{n} \sum_{k=1}^{n_d} \left[\sum_{k_2=1}^{n_d} c_{q,k_2} \left(b_{k_2}^{\prime} \circ \phi \cdot \phi^{\prime} \right)\left(\sum_{p_1=1}^{n}\sum_{k_1=1}^{n_d}a_{p_1,q,k_1}b_{k_1}\left(x_{i,p_1}\right)\right) \cdot b_{k}(x_{i,p})\right]\\
        & \cdot \left[\sum_{k_2=1}^{n_d} c_{q,k_2} \left(b_{k_2}^{\prime} \circ \phi \cdot \phi^{\prime} \right)\left(\sum_{p_1=1}^{n}\sum_{k_1=1}^{n_d}a_{p_1,q,k_1}b_{k_1}\left(x_{j,p_1}\right)\right) \cdot b_{k}(x_{j,p})\right],
    \end{split}
\end{equation*}
and 
\begin{equation*}
    \begin{split}
        & \left\langle \frac{\partial s_{i}}{\partial \bc}, \frac{\partial s_{j}}{\partial \bc}\right \rangle \\
        = & \sum_{q=1}^{m} \sum_{k=1}^{n_d} \frac{\partial s_i}{\partial c_{q,k}} \cdot \frac{\partial s_j}{\partial c_{q,k}}\\
        = & \frac{1}{mN} \sum_{q=1}^{m} \sum_{k=1}^{n_d} b_{k} \left( \phi\left(\sum_{p_1=1}^{n}\sum_{k_1=1}^{n_d}a_{p_1,q,k_1}b_{k_1}\left(x_{i,p_1}\right)\right) \right) \cdot  b_{k} \left( \phi\left(\sum_{p_1=1}^{n}\sum_{k_1=1}^{n_d}a_{p_1,q,k_1}b_{k_1}\left(x_{j,p_1}\right)\right) \right).
    \end{split}
\end{equation*}
Therefore, we can regard $G_{i,j}$ as the summation of $m$ independent and identically distributed random variables, leading to the independence of $\bG^{\infty}$ on $m$ after taking the expectation.

\subsection{Proof for \Cref{lemma_init_loss}}
\label{appendix_proof_lemma_init_loss}
Recall the definition of the loss function
\begin{equation*}
    \mathcal{L}(0) = \frac{1}{N}\sum_{i=1}^{N} \left( f_{i}(\ba(0),\bc(0)) - y_i\right)^2 \leq \frac{2}{N} \sum_{i=1}^{N} y_i^2 + \frac{2}{N} \sum_{i=1}^{N} f_{i}(\ba(0),\bc(0))^2.
\end{equation*}
Note that 
\begin{equation*}
    f_{i}(\ba(0),\bc(0)) = \frac{1}{\sqrt{m}}\sum_{q=1}^{m} \sum_{k=1}^{n_d}c_{q,k}(0) b_k\left( \phi\left(\sum_{p_1=1}^{n} \sum_{k_1=1}^{n_d} a_{p_1,q,k_1}(0) b_{k_1}\left(x_{i,p} \right) \right)\right),
\end{equation*}
where $b_k \circ \phi$ is bounded  due to \Cref{assump_bounded} and $c_{q,k}(0)$ are independent.  According to \cite[Theorem 3.1]{kuchibhotla2022moving} with $\alpha=2$ and $a_i = \frac{1}{\sqrt{m n_d}}$, with probability of at least $1-2e^{-M}$, we have
\begin{equation*}
    \left|f_i(\ba(0),\bc(0)) \right| \lesssim \sqrt{n_d M}.
\end{equation*}
Let $\delta = 2N e^{-M}$, we have $M = \log \frac{2N}{\delta}$, and therefore,
\begin{equation*}
    \mathcal{L}(0) \lesssim n_d \cdot \left(\log \frac{N}{\delta} \right).
\end{equation*}

\subsection{Proof for \Cref{lemma_init_gram}}
\label{appendix_proof_lemma_init_gram}
For each $q \in [m]$, $\left\|\bc_{q}\right\|_2^2 = \sum_{k=1}^{n_d} c_{q,k}^2$ follows the chi-squared distribution with $n_d$ degrees of freedom. Therefore, with probability of at least $1-\frac{\delta}{2m}$ over the initialization, we have
\begin{equation}
\label{eq_appx_3}
    \left\|\bc_{q}\right\|_2 \lesssim \left(1+ \sqrt{\frac{2}{n_d} \log \frac{2m}{\delta}}\right) \cdot \sqrt{n_d} \lesssim \sqrt{n_d} + \sqrt{\log \frac{m}{\delta}}.
\end{equation}
Let $\sqrt{n_d} + \sqrt{\log \frac{m}{\delta}} \simeq M_c$, then with probability of at least $1- \frac{\delta}{2}$ over the initialization, we have 
\begin{equation*}
    \left\|\bc_{q}\right\|_2 \leq \frac{1}{2}M_c,
\end{equation*}
for all $q \in [m]$.
Note that $G_{i,j} = \left\langle \frac{\partial s_{i}}{\partial \ba}, \frac{\partial s_{j}}{\partial \ba}\right \rangle + \left\langle \frac{\partial s_{i}}{\partial \bc}, \frac{\partial s_{j}}{\partial \bc}\right \rangle := S_{i,j} + Q_{i,j}$, where we denote $S_{i,j}$ and $Q_{i,j}$ as $\left\langle \frac{\partial s_{i}}{\partial \ba}, \frac{\partial s_{j}}{\partial \ba}\right \rangle $ and $\left\langle \frac{\partial s_{i}}{\partial \bc}, \frac{\partial s_{j}}{\partial \bc}\right \rangle$, respectively.
\begin{equation*}
    \begin{split}
        S_{i,j} = & \left\langle \frac{\partial s_{i}}{\partial \ba}, \frac{\partial s_{j}}{\partial \ba}\right \rangle \\
        = & \sum_{q=1}^{m} \sum_{p=1}^{n} \sum_{k=1}^{n_d} \frac{\partial s_i}{\partial a_{p,q,k}} \cdot \frac{\partial s_j}{\partial a_{p,q,k}} \\
        = & \frac{1}{mN} \sum_{q=1}^{m} \sum_{p=1}^{n} \sum_{k=1}^{n_d} \left[\sum_{k_2=1}^{n_d} c_{q,k_2} \left(b_{k_2}^{\prime} \circ \phi \cdot \phi^{\prime} \right)\left(\sum_{p_1=1}^{n}\sum_{k_1=1}^{n_d}a_{p_1,q,k_1}b_{k_1}\left(x_{i,p_1}\right)\right) \cdot b_{k}(x_{i,p})\right]\\
        & \cdot \left[\sum_{k_2=1}^{n_d} c_{q,k_2} \left(b_{k_2}^{\prime} \circ \phi \cdot \phi^{\prime} \right)\left(\sum_{p_1=1}^{n}\sum_{k_1=1}^{n_d}a_{p_1,q,k_1}b_{k_1}\left(x_{j,p_1}\right)\right) \cdot b_{k}(x_{j,p})\right]\\
        := & \frac{1}{m} \sum_{q=1}^{m} X_{q},
    \end{split}
\end{equation*}
where $X_q$ are independent and identically distributed random variables, and 
\begin{equation}
\label{eq_appx_7}
    \left|X_q\right| \lesssim \frac{n_d^2 n}{N} \qquad\left\| \bc_{q}\right\|_2^2 \lesssim \frac{n_d^2 n}{N} M_c^2,
\end{equation}
under the assumption of the boundedness of the transformation function and basis functions (and their derivatives) in \Cref{assump_bounded}.
The Hoeffding's inequality shows that 
\begin{equation}
\label{eq_appx_8}
    \begin{split}
        & \mathbb{P}\left(\left|S_{i,j}(0) - \mathbb{E}\left[S_{i,j} \right]\right| > \epsilon \right) \\
        = & \mathbb{P}\left(\left|\frac{1}{m} \sum_{q=1}^{m} X_{q} - \mathbb{E}\left[X_{q} \right]\right| > \epsilon \right)\\
        \leq & 2 \exp\left(-\frac{m \epsilon^2 N^2}{2 n_d^4 n^2 M_c^4}\right).
    \end{split}
\end{equation}
Let $\epsilon = \frac{\sigma_{\min}}{8N}$ and the last term in the above inequality to be less than $\frac{\delta}{4N^2}$, then with probability of at least $1-\frac{\delta}{4N^2}$, we have
\begin{equation*}
    \left|S_{i,j}(0) - \mathbb{E}\left[S_{i,j} \right]\right| \leq \frac{\sigma_{\min}}{16N},
\end{equation*}
under the condition that
\begin{equation*}
    m \gtrsim \frac{n_d^4 n^2}{\sigma_{\min}^2} \log \left(\frac{N}{\delta}\right) \cdot \left(n_d^2 + \left(\log \frac{m}{\delta}\right)^2\right).
\end{equation*}
Therefore, with probability of at least $1-\frac{\delta}{4}$, 
\begin{equation*}
    \left|S_{i,j}(0) - \mathbb{E}\left[S_{i,j} \right]\right| \leq \frac{\sigma_{\min}}{16N},
\end{equation*}
holds for all $i,j \in [N]$. 

\begin{equation*}
    \begin{split}
        Q_{i,j} = & \left\langle \frac{\partial s_{i}}{\partial \bc}, \frac{\partial s_{j}}{\partial \bc}\right \rangle \\
        = & \sum_{q=1}^{m} \sum_{k=1}^{n_d} \frac{\partial s_i}{\partial c_{q,k}} \cdot \frac{\partial s_j}{\partial c_{q,k}}\\
        = & \frac{1}{mN} \sum_{q=1}^{m} \sum_{k=1}^{n_d} b_{k} \left( \phi\left(\sum_{p_1=1}^{n}\sum_{k_1=1}^{n_d}a_{p_1,q,k_1}b_{k_1}\left(x_{i,p_1}\right)\right) \right) \cdot  b_{k} \left( \phi\left(\sum_{p_1=1}^{n}\sum_{k_1=1}^{n_d}a_{p_1,q,k_1}b_{k_1}\left(x_{j,p_1}\right)\right) \right)\\
        := & \frac{1}{m} \sum_{q=1}^{m} Y_q,
    \end{split}
\end{equation*}
where $Y_q$ are independent and identically distributed, and 
\begin{equation*}
    \left|Y_q\right| \leq \frac{n_d}{N},
\end{equation*}
under the assumption of the boundedness of the transformation function and basis functions (and their derivatives) in \Cref{assump_bounded}.
The Hoeffding's inequality shows that 
\begin{equation*}
    \begin{split}
        & \mathbb{P}\left(\left|Q_{i,j}(0) - \mathbb{E}\left[Q_{i,j} \right]\right| > \epsilon \right) \\
        = & \mathbb{P}\left(\left|\frac{1}{m} \sum_{q=1}^{m} Y_{q} - \mathbb{E}\left[Y_{q} \right]\right| > \epsilon \right)\\
        \leq & 2 \exp\left(-\frac{m \epsilon^2 N^2}{2 n_d^2}\right).
    \end{split}
\end{equation*}
Let $\epsilon = \frac{\sigma_{\min}}{8N}$ and the last term in the above inequality to be less than $\frac{\delta}{4N^2}$, then with probability of at least $1-\frac{\delta}{4N^2}$, we have
\begin{equation*}
    \left|Q_{i,j}(0) - \mathbb{E}\left[Q_{i,j} \right]\right| \leq \frac{\sigma_{\min}}{16N},
\end{equation*}
under the condition that
\begin{equation*}
    m \gtrsim \frac{n_d^2}{\sigma_{\min}^2} \log \left(\frac{N}{\delta}\right).
\end{equation*}
Therefore, with probability of at least $1-\frac{\delta}{4}$, 
\begin{equation*}
    \left|S_{i,j}(0) - \mathbb{E}\left[S_{i,j} \right]\right| \leq \frac{\sigma_{\min}}{16N},
\end{equation*}
holds for all $i,j \in [N]$. Summarizing the above analysis, we have
\begin{equation*}
    \left\|\bG(0) - \bG^{\infty} \right\|_2 \leq \left\|\bG(0) - \bG^{\infty} \right\|_F \leq N \cdot \left(\frac{\sigma_{\min}}{16N} + \frac{\sigma_{\min}}{16N} \right) \leq \frac{\sigma_{\min}}{8},
\end{equation*}
with probability of at least $1-\delta$, under the condition that
\begin{equation*}
    m \gtrsim \frac{n_d^4 n^2}{\sigma_{\min}^2} \log \left(\frac{N}{\delta}\right) \cdot \left(n_d + \log \frac{m}{\delta}\right).
\end{equation*}

\subsection{Proof for \Cref{lemma_local}}
\label{appendix_proof_lemma_local}
Recall that 
\begin{equation*}
    \begin{split}
        S_{i,j} = & \left\langle \frac{\partial s_{i}}{\partial \ba}, \frac{\partial s_{j}}{\partial \ba}\right \rangle \\
        = & \sum_{q=1}^{m} \sum_{p=1}^{n} \sum_{k=1}^{n_d} \frac{\partial s_i}{\partial a_{p,q,k}} \cdot \frac{\partial s_j}{\partial a_{p,q,k}} \\
        = & \frac{1}{mN} \sum_{q=1}^{m} \sum_{p=1}^{n} \sum_{k=1}^{n_d} \left[\sum_{k_2=1}^{n_d} c_{q,k_2} \left(b_{k_2}^{\prime} \circ \phi \cdot \phi^{\prime} \right)\left(\sum_{p_1=1}^{n}\sum_{k_1=1}^{n_d}a_{p_1,q,k_1}b_{k_1}\left(x_{i,p_1}\right)\right) \cdot b_{k}(x_{i,p})\right]\\
        & \cdot \left[\sum_{k_2=1}^{n_d} c_{q,k_2} \left(b_{k_2}^{\prime} \circ \phi \cdot \phi^{\prime} \right)\left(\sum_{p_1=1}^{n}\sum_{k_1=1}^{n_d}a_{p_1,q,k_1}b_{k_1}\left(x_{j,p_1}\right)\right) \cdot b_{k}(x_{j,p})\right],
    \end{split}
\end{equation*}
then
\begin{equation*}
    \begin{split}
        &\frac{\partial S_{i,j}}{\partial a_{p,q,k}}\\
        = & \frac{1}{m N} \sum_{p_2=1}^{n} \sum_{k_3=1}^{n_d} \left[\sum_{k_2=1}^{n_d} c_{q,k_2} \left(b_{k_2}^{\prime} \circ \phi \cdot \phi^{\prime} \right)\left(\sum_{p_1=1}^{n}\sum_{k_1=1}^{n_d}a_{p_1,q,k_1}b_{k_1}\left(x_{i,p_1}\right)\right) \cdot b_{k_3}(x_{i,p_2})\right] \\
        & \cdot \left[\sum_{k_2=1}^{n_d} c_{q,k_2} \left(b_{k_2}^{\prime \prime} \circ \phi \cdot \phi^{\prime 2} \right)\left(\sum_{p_1=1}^{n}\sum_{k_1=1}^{n_d}a_{p_1,q,k_1}b_{k_1}\left(x_{j,p_1}\right)\right) \cdot b_{k_3}(x_{j,p}) b_{k}(x_{j,p_2})\right.\\
        & \left. + c_{q,k_2} \left(b_{k_2}^{\prime} \circ \phi \cdot \phi^{\prime \prime} \right)\left(\sum_{p_1=1}^{n}\sum_{k_1=1}^{n_d}a_{p_1,q,k_1}b_{k_1}\left(x_{j,p_1}\right)\right) \cdot b_{k_3}(x_{j,p}) b_{k}(x_{j,p_2})\right] \\
        & + \left[\sum_{k_2=1}^{n_d} c_{q,k_2} \left(b_{k_2}^{\prime \prime} \circ \phi \cdot \phi^{\prime 2} \right)\left(\sum_{p_1=1}^{n}\sum_{k_1=1}^{n_d}a_{p_1,q,k_1}b_{k_1}\left(x_{i,p_1}\right)\right) \cdot b_{k_3}(x_{i,p}) b_{k}(x_{i,p_2})\right.\\
        & \left. + c_{q,k_2} \left(b_{k_2}^{\prime} \circ \phi \cdot \phi^{\prime \prime} \right)\left(\sum_{p_1=1}^{n}\sum_{k_1=1}^{n_d}a_{p_1,q,k_1}b_{k_1}\left(x_{i,p_1}\right)\right) \cdot b_{k_3}(x_{i,p}) b_{k}(x_{i,p_2})\right] \\
        & \cdot \left[\sum_{k_2=1}^{n_d} c_{q,k_2} \left(b_{k_2}^{\prime} \circ \phi  \cdot \phi^{\prime} \right) \left(\sum_{p_1=1}^{n}\sum_{k_1=1}^{n_d}a_{p_1,q,k_1}b_{k_1}\left(x_{j,p_1}\right)\right) \cdot b_{k_3}(x_{j,p_2})\right].
    \end{split}
\end{equation*}
After carefully rearranging the above equality, we obtain
\begin{equation}
\label{eq_appx_9}
    \left| \frac{\partial S_{i,j}}{\partial a_{p,q,k}} \right| \lesssim \frac{n_d^{2} n }{mN} \left\|\bc_{q} \right\|_2^2,
\end{equation}
and thus,
\begin{equation}
\label{eq_appx_10}
    \left\| \frac{\partial S_{i,j}}{\partial \ba_{q}}\right\|_2 \lesssim \frac{n_d^{5/2} n^{3/2}}{mN} \left\|\bc_{q} \right\|_2^2.
\end{equation}

\begin{equation*}
    \begin{split}
        & \frac{\partial S_{i,j}}{\partial c_{q,k}} \\
        = & \frac{1}{m N} \sum_{p_2=1}^{n} \sum_{k_3=1}^{n_d} \left\{\left[\sum_{k_2=1}^{n_d} c_{q,k_2} \left(b_{k_2}^{\prime} \circ \phi \cdot \phi^{\prime} \right)\left(\sum_{p_1=1}^{n}\sum_{k_1=1}^{n_d}a_{p_1,q,k_1}b_{k_1}\left(x_{i,p_1}\right)\right) \cdot b_{k_3}(x_{i,p_2})\right] \right.\\
        & \cdot \left(b_{k}^{\prime} \circ \phi \cdot \phi^{\prime} \right)\left(\sum_{p_1=1}^{n}\sum_{k_1=1}^{n_d}a_{p_1,q,k_1}b_{k_1}\left(x_{j,p_1}\right)\right) \cdot b_{k_3}(x_{j,p_2}) \\
        & + \left(b_{k}^{\prime} \circ \phi \cdot \phi^{\prime} \right)\left(\sum_{p_1=1}^{n}\sum_{k_1=1}^{n_d}a_{p_1,q,k_1}b_{k_1}\left(x_{i,p_1}\right)\right) \cdot b_{k_3}(x_{i,p_2}) \\
        & \cdot \left.\left[\sum_{k_2=1}^{n_d} c_{q,k_2} \left(b_{k_2}^{\prime} \circ \phi  \cdot \phi^{\prime} \right) \left(\sum_{p_1=1}^{n}\sum_{k_1=1}^{n_d}a_{p_1,q,k_1}b_{k_1}\left(x_{j,p_1}\right)\right) \cdot b_{k_3}(x_{j,p_2})\right] \right\}.
    \end{split}
\end{equation*}
Rearranging the above equality, we have
\begin{equation}
\label{eq_appx_11}
    \left| \frac{\partial S_{i,j}}{\partial c_{q,k}}\right| \lesssim \frac{n_d^{3/2} n}{mN} \left\|\bc_{q} \right\|_2,
\end{equation}
and thus
\begin{equation}
\label{eq_appx_12}
    \left\| \frac{\partial S_{i,j}}{\partial \bc_{q}}\right\|_2 \leq \frac{n_d^2 n}{mN} \left\|\bc_{q} \right\|_2.
\end{equation}

Similarly, note that
\begin{equation*}
    \begin{split}
        Q_{i,j} = & \left\langle \frac{\partial s_{i}}{\partial \bc}, \frac{\partial s_{j}}{\partial \bc}\right \rangle \\
        = & \sum_{q=1}^{m} \sum_{k=1}^{n_d} \frac{\partial s_i}{\partial c_{q,k}} \cdot \frac{\partial s_j}{\partial c_{q,k}}\\
        = & \frac{1}{mN} \sum_{q=1}^{m} \sum_{k_2=1}^{n_d} b_{k_2} \left( \phi\left(\sum_{p_1=1}^{n}\sum_{k_1=1}^{n_d}a_{p_1,q,k_1}b_{k_1}\left(x_{i,p_1}\right)\right) \right) \cdot  b_{k_2} \left( \phi\left(\sum_{p_1=1}^{n}\sum_{k_1=1}^{n_d}a_{p_1,q,k_1}b_{k_1}\left(x_{j,p_1}\right)\right) \right),
    \end{split}
\end{equation*}
we have 
\begin{equation*}
    \begin{split}
    & \frac{\partial Q_{i,j}}{\partial a_{p,q,k}}\\
    = & \frac{1}{mN}\sum_{k_2=1}^{n_d} \left(b_{k_2}^{\prime} \circ \phi  \cdot \phi^{\prime} \right)\left( \sum_{p_1=1}^{n}\sum_{k_1=1}^{n_d}a_{p_1,q,k_1}b_{k_1}\left(x_{i,p_1}\right)\right) \cdot b_{k}(x_{i,p}) \cdot b_{k_2}  \left(\phi\left( \sum_{p_1=1}^{n}\sum_{k_1=1}^{n_d}a_{p_1,q,k_1}b_{k_1}\left(x_{j,p_1}\right)\right)\right) \\
    & + b_{k_2}  \left(\phi\left( \sum_{p_1=1}^{n}\sum_{k_1=1}^{n_d}a_{p_1,q,k_1}b_{k_1}\left(x_{i,p_1}\right)\right)\right) \cdot \left(b_{k_2}^{\prime} \circ \phi  \cdot \phi^{\prime} \right)\left( \sum_{p_1=1}^{n}\sum_{k_1=1}^{n_d}a_{p_1,q,k_1}b_{k_1}\left(x_{j,p_1}\right)\right) \cdot b_{k}(x_{j,p}), 
    \end{split}
\end{equation*}
and
\begin{equation*}
    \frac{\partial Q_{i,j}}{\partial c_{q,k}} = 0.
\end{equation*}
Therefore, we obtain that
\begin{equation*}
    \left| \frac{\partial Q_{i,j}}{\partial a_{p,q,k}}\right| \lesssim \frac{n_d}{mN},
\end{equation*}
and thus 
\begin{equation*}
    \left\| \frac{\partial Q_{i,j}}{\partial \ba_{q}}\right\|_2 \lesssim \frac{n_d^{3/2} n^{1/2}}{mN},\quad\text{and}\quad\left\| \frac{\partial Q_{i,j}}{\partial \bc_{q}}\right\|_2 = 0.
\end{equation*}
Since $G_{i,j}=S_{i,j}+Q_{i,j}$, we have
\begin{equation*}
    \left\| \frac{\partial G_{i,j}}{\partial \ba_{q}}\right\|_2 \lesssim \frac{n_d^{5/2} n^{3/2}}{mN} \left\|\bc_{q} \right\|_2^2,\quad\text{and}\quad\left\| \frac{\partial G_{i,j}}{\partial \bc_{q}}\right\|_2 \leq \frac{n_d^2 n}{mN} \left\|\bc_{q} \right\|_2.
\end{equation*}
Summarizing the above results, we have
\begin{equation}
\label{eq_appx_2}
    \begin{split}
        & \left| G_{i,j}(\bm{a},\bm{c}) - G_{i,j}(\bm{a}(0),\bm{c}(0)) \right|\\
        \leq & \left|\sum_{q=1}^{m}\left\langle\frac{\partial G_{i,j}}{\partial \ba_q},\ba_{q} - \ba_{q}(0) \right\rangle + \left\langle \frac{\partial G_{i,j}}{\partial \bc_q},\bc_{q} - \bc_{q}(0)\right\rangle\right|\\
        \leq & \sum_{q=1}^{m} \left\|\frac{\partial G_{i,j}}{\partial \ba_q} \right\|_2 \cdot \left\| \ba_{q} - \ba_{q}(0)\right\|_2 + \left\|\frac{\partial G_{i,j}}{\partial \bc_q} \right\|_2 \cdot \left\| \bc_{q} - \bc_{q}(0)\right\|_2\\
        \lesssim & \sum_{q=1}^{m} \frac{n_d^{5/2} n^{3/2}}{mN} \left\|\bc_{q} \right\|_2^2 \cdot \left\| \ba_{q} - \ba_{q}(0)\right\|_2 + \frac{n_d^2 n}{mN} \left\|\bc_{q} \right\|_2 \cdot \left\| \bc_{q} - \bc_{q}(0)\right\|_2 \\
        \lesssim &  \frac{n_d^{5/2} n^{3/2}}{\sqrt{m}N} M_c^2 \cdot R_a + \frac{n_d^2 n}{\sqrt{m} N} M_c \cdot R_c.
    \end{split}
\end{equation}
Let 
\begin{equation*}
    \frac{n_d^{5/2} n^{3/2}}{\sqrt{m}N} M_c^2 \cdot R_a + \frac{n_d^2 n}{\sqrt{m} N} M_c \cdot R_c \lesssim \frac{\sigma_{\min}}{8N},
\end{equation*}
such that $\left| G_{i,j}(\bm{a},\bm{c}) - G_{i,j}(\bm{a}(0),\bm{c}(0)) \right| \leq \frac{\sigma_{\min}}{8N}$, we equivalently require that 
\begin{equation*}
    R_a \lesssim \frac{\sigma_{\min} \sqrt{m}}{n_d^{5/2} n^{3/2} M_c^2},\quad\text{and}\quad R_c \lesssim \frac{\sigma_{\min}\sqrt{m}}{n_d^2 n M_c}.
\end{equation*}
Therefore, 
\begin{equation*}
    \left\|\bG(\ba,\bc) - \bG(\ba(0),\bc(0)) \right\|_2 \leq N \cdot \frac{\sigma_{\min}}{8N} \leq \frac{\sigma_{\min}}{8}.
\end{equation*}

\subsection{Proof for \Cref{lemma_next_step_inball}}
\label{appendix_proof_lemma_next_step_inball}
Recall that
\begin{equation*}
    \frac{\partial f_i}{\partial a_{p,q,k}} = \frac{1}{\sqrt{m}} \sum_{k_2=1}^{n_d} c_{q,k_2} \left( b_{k_2}^{\prime} \circ \phi \cdot \phi^{\prime} \right) \left( \sum_{p_1=1}^{n}\sum_{k_1=1}^{n_d}a_{p_1,q,k_1}b_{k_1}\left(x_{i,p_1}\right)\right) \cdot b_{k}(x_{i,p}),
\end{equation*}
and
\begin{equation*}
    \frac{\partial f_i}{\partial c_{q,k}} = \frac{1}{\sqrt{m}} b_{k}\left(\phi\left(\sum_{p_1=1}^{n}\sum_{k_1=1}^{n_d}a_{p_1,q,k_1}b_{k_1}\left(x_{i,p_1}\right)\right)\right),
\end{equation*}
then 
\begin{equation*}
    \left\|\frac{\partial f_i}{\partial \ba_{q}}\right\|_2 \lesssim \sqrt{\frac{n_d}{m}} \left\| \bc_q\right\|_2,\quad \text{and}\quad \left\| \frac{\partial f_i}{\partial \bc_{q}}\right\|_2 \lesssim \sqrt{\frac{n_d}{m}}.
\end{equation*}
By the definition of $\bG^{\infty}$,
\begin{equation}
\label{eq_appx_14}
    G_{i,j}^{\infty} = \mathbb{E}\left[G_{i,j}\right] \lesssim \mathbb{E}\left[\frac{1}{m N} \sum_{q=1}^{m} \left(\left\| \bc_q\right\|_2^2 + 1\right)\right] \lesssim \frac{n_d}{N},
\end{equation}
which implies that $\sigma_{\max} \lesssim n_d$.
We have
\begin{equation*}
    \begin{split}
        \left\|\ba(t+1) - \ba(0) \right\|_2 \leq &  \sum_{\tau=0}^{t}\left\|\ba(\tau+1) - \ba(\tau) \right\|_2 \\
        \leq & \eta \sum_{\tau=0}^{t} \left\| \frac{\partial \mathcal{L}(\tau)}{\partial \ba}\right\|_2 \\
        \leq & \eta \sum_{\tau=0}^{t} \left\| \left[\begin{array}{ccc}
            \frac{\partial \bs_1(\tau)}{\partial \ba} & \cdots & \frac{\partial \bs_N(\tau)}{\partial \ba}
        \end{array}\right]\right\|_2 \cdot \left\| \bs(\tau)\right\|_2\\
        \lesssim & \eta \sqrt{\sigma_{\max}} \sum_{\tau=0}^{t} \left\| \bs(\tau)\right\|_2\\
        \lesssim & \eta \sqrt{\sigma_{\max}}  \sum_{\tau=0}^{t} \left(1- \eta \cdot \frac{\sigma_{\min}}{2} \right)^{\tau} \cdot \left\| \bs(0)\right\|_2\\
        \lesssim & \frac{\sqrt{\sigma_{\max}} }{\sigma_{\min}} \cdot \left\| \bs(0)\right\|_2 \lesssim \frac{\sqrt{n_d} }{\sigma_{\min}} \cdot \left\| \bs(0)\right\|_2,
    \end{split}
\end{equation*}
and
\begin{equation*}
    \begin{split}
        \left\|\bc_q(t+1) - \bc_q(0) \right\|_2 \leq & \eta \sum_{\tau=0}^{t}\left\|\bc_q(\tau+1) - \bc_q(\tau) \right\|_2 \\
        \leq & \eta \sum_{\tau=0}^{t} \left\| \frac{\partial \mathcal{L}(\tau)}{\partial \bc_q}\right\|_2 \\
        \leq & \eta \sum_{\tau=0}^{t} \left( \sum_{i=1}^{N}\left|s_i(\tau)\right| \cdot \left\| \frac{\partial s_i(\tau)}{\partial \bc_q}\right\|_2\right)\\
        \lesssim & \eta \sqrt{\frac{n_d}{m}} \sum_{\tau=0}^{t} \left\| \bs(\tau)\right\|_2\\
        \lesssim & \eta \sqrt{\frac{n_d}{m}} \sum_{\tau=0}^{t} \left(1- \eta \cdot \frac{\sigma_{\min}}{2} \right)^{\tau} \cdot \left\| \bs(0)\right\|_2\\
        \lesssim & \frac{\sqrt{n_d}}{\sqrt{m} \sigma_{\min}} \cdot \left\| \bs(0)\right\|_2,
    \end{split}
\end{equation*}
which implies that
\begin{equation*}
     \left\|\bc(t+1) - \bc(0) \right\|_2 \lesssim \frac{\sqrt{n_d}}{ \sigma_{\min}} \cdot \left\| \bs(0)\right\|_2.
\end{equation*}
Let
\begin{equation*}
    \frac{\sqrt{n_d} }{\sigma_{\min}} \cdot \left\| \bs(0)\right\|_2 \lesssim R_a,~\frac{\sqrt{n_d}}{\sigma_{\min}} \cdot \left\| \bs(0)\right\|_2\lesssim R_c,~\text{and}~\frac{\sqrt{n_d}}{\sqrt{m} \sigma_{\min}} \cdot \left\| \bs(0)\right\|_2 \lesssim M_c,
\end{equation*}
such that
\begin{equation*}
    \left\|\ba(t+1) - \ba(0) \right\|_2 \leq R_a,~\left\|\bc(t+1) - \bc(0) \right\|_2 \leq R_c,~\text{and}~\left\|\bc_{q}(t+1)\right\|_2 \leq M_c,
\end{equation*}
we require
\begin{equation*}
    m \gtrsim \frac{n_d^{7} n^3}{\sigma_{\min}^{4}} \left(n_d^2 + \left(\log \frac{m}{\delta} \right)^2 \right) \cdot \left(\log \frac{N}{\delta}\right),
\end{equation*}
where we plug the results on $\mathcal{L}(0)$ from \Cref{lemma_init_loss} into the above inequality.

\subsection{Proof for \Cref{lemma_bound_tau_res}}
\label{appendix_proof_lemma_bound_tau_res}
We consider the second-order derivative of $s_i$ with respect to $\ba$ and $\bc$ to bound the error term $\chi_i$. Recall that
\begin{equation*}
    \frac{\partial s_i}{\partial a_{p,q,k}} = \frac{1}{\sqrt{mN}} \sum_{k_2=1}^{n_d} c_{q,k_2} \left( b_{k_2}^{\prime} \circ \phi \cdot \phi^{\prime} \right) \left( \sum_{p_1=1}^{n}\sum_{k_1=1}^{n_d}a_{p_1,q,k_1}b_{k_1}\left(x_{i,p_1}\right)\right) \cdot b_{k}(x_{i,p}),
\end{equation*}
then we have
\begin{equation*}
    \begin{split}
        & \frac{\partial^2 s_i}{\partial a_{p_0,q_0,k_0} \partial a_{p,q,k}}\\
        = &  \frac{1}{\sqrt{mN}} \sum_{k_2=1}^{n_d} c_{q,k_2} \left(b_{k_2}^{\prime \prime} \circ \phi \cdot \phi^{\prime 2}\right)\left( \sum_{p_1=1}^{n}\sum_{k_1=1}^{n_d}a_{p_1,q,k_1}b_{k_1}\left(x_{i,p_1}\right)\right) b_{k}(x_{i,p}) b_{k_0}(x_{i,p_0})\\
        & + c_{q,k_2} \left(b_{k_2}^{\prime} \circ \phi \cdot \phi^{\prime \prime}\right)\left( \sum_{p_1=1}^{n}\sum_{k_1=1}^{n_d}a_{p_1,q,k_1}b_{k_1}\left(x_{i,p_1}\right)\right) b_{k}(x_{i,p}) b_{k_0}(x_{i,p_0}),
    \end{split}
\end{equation*}
\begin{equation*}
    \begin{split}
        & \frac{\partial^2 s_i}{\partial c_{q_0,k_0} \partial a_{p,q,k}}\\
        = & \frac{1}{\sqrt{mN}} \left(b_{k_0}^{\prime} \circ \phi \cdot \phi^{\prime} \right)\left( \sum_{p_1=1}^{n}\sum_{k_1=1}^{n_d}a_{p_1,q,k_1}b_{k_1}\left(x_{i,p_1}\right)\right) b_{k}(x_{i,p}),
    \end{split}
\end{equation*}
and
\begin{equation*}
    \frac{\partial^2 s_i}{\partial c_{q_0,k_0} \partial c_{q,k}} = 0.
\end{equation*}
Summarizing the above results, we conclude that
\begin{equation*}
    \left|\frac{\partial^2 s_i}{\partial a_{p_0,q_0,k_0} \partial a_{p,q,k}}\right| \lesssim \frac{1}{\sqrt{mN}} \left\| \bc_{q}\right\|_2,\quad\text{and}\quad \left|\frac{\partial^2 s_i}{\partial c_{q_0,k_0} \partial a_{p,q,k}}\right| \lesssim \frac{1}{\sqrt{mN}}.
\end{equation*}
Recall that 
\begin{equation*}
    \begin{split}
        & \left\| \ba(\tau+1) - \ba(\tau)\right\|_2^2 \\
        = & \eta^2 \left\| \frac{\partial \mathcal{L}(\tau)}{\partial \ba}\right\|_2^2\\
        = & \eta^2 \left(\sum_{i=1}^{N} s_i(\tau) \frac{\partial s_i(\tau)}{\partial \ba_q} \right)^2\\
        \leq & \eta^2 \left\| \left[\begin{array}{ccc}
            \frac{\partial \bs_1(\tau)}{\partial \ba} & \cdots & \frac{\partial \bs_N(\tau)}{\partial \ba}
        \end{array}\right]\right\|_2^2 \cdot \left\| \bs(\tau)\right\|_2^2\\
        \lesssim & \eta^2 \sigma_{\max} \left\| \bs(\tau)\right\|_2^2 \lesssim \eta^2 n_d \left\| \bs(\tau)\right\|_2^2,
    \end{split}
\end{equation*}
and
\begin{equation*}
    \begin{split}
        & \left\| \bc_q(\tau+1) - \bc_q(\tau)\right\|_2^2 \\
        = & \eta^2 \left\| \frac{\partial \mathcal{L}(\tau)}{\partial \bc_q}\right\|_2^2\\
        = & \eta^2 \left(\sum_{i=1}^{N} s_i(\tau) \frac{\partial s_i(\tau)}{\partial \bc_q} \right)^2\\
        \leq & \frac{\eta^2 n_d}{m} \left\| \bs(\tau)\right\|_2^2.
    \end{split}
\end{equation*}
Therefore, we have
\begin{equation}\label{eq_appx_4}
\begin{split}
    & \left|\chi_i(\tau)\right|\\
    \leq & \frac{1}{2}\left\|\frac{\partial^2 s_i}{\partial \ba \partial \ba} \right\|_2 \cdot \left\| \ba(\tau+1) - \ba(\tau)\right\|_2^2 + \left\|\frac{\partial^2 s_i}{\partial \bc \partial \ba} \right\|_2 \cdot \left\| \ba(\tau+1) - \ba(\tau)\right\|_2 \left\| \bc(\tau+1) - \bc(\tau)\right\|_2\\
    \lesssim & \sum_{q=1}^{m} \sqrt{\frac{n_d n}{m N}} \left\|\bc_{q} \right\|_2 \cdot \left\| \ba_q(\tau+1) - \ba_q(\tau)\right\|_2^2 + \sqrt{\frac{1}{mN}} \cdot \left( \left\| \ba_q(\tau+1) - \ba_q(\tau)\right\|_2^2 + \left\| \bc_q(\tau+1) - \bc_q(\tau)\right\|_2^2\right)\\
    \lesssim & \eta^2 \frac{n_d^{3/2} \sqrt{n}}{\sqrt{mN}} M_c \mathcal{L}(\tau),
\end{split}
\end{equation}
which implies 
\begin{equation*}
    \left\|\bm{\chi}(\tau) \right\|_2 \lesssim \eta^2 \frac{n_d^{3/2} \sqrt{n}}{\sqrt{m}} M_c \mathcal{L}(\tau).
\end{equation*}
Let 
\begin{equation*}
    \eta^2 \frac{n_d^{3/2} \sqrt{n}}{\sqrt{m}} M_c \mathcal{L}(\tau) \lesssim \eta \sigma_{\min} \left\| \bs(\tau)\right\|_2,
\end{equation*}
such that
\begin{equation*}
    \left\|\bm{\chi}(\tau) \right\|_2 \leq \frac{\eta \sigma_{\min}}{4} \left\| \bs(\tau)\right\|_2,
\end{equation*}
under the condition that
\begin{equation*}
    m \gtrsim \frac{\eta^2 n_d^3 n M_c^2}{\sigma_{\min}^2} \mathcal{L}(0),
\end{equation*}
which can be guaranteed by the condition in \Cref{lemma_next_step_inball}.

\subsection{Proof for \Cref{theorem_gd}}
\label{appendix_proof_theorem_gd}
Rearranging Lemmas \ref{lemma_init_loss} to \ref{lemma_bound_tau_res}, we have
\begin{equation*}
    \begin{split}
        \left\|\bs(t+1) \right\|_2 = & \left\|\bs(t+1) -\bs(t)+ \bs(t)\right\|_2 \\
        \leq & \left\| \left \langle \frac{\partial \bs(\tau)}{\partial \ba}, \ba(\tau+1)-\ba(\tau)\right\rangle + \left \langle \frac{\partial \bs(\tau)}{\partial \bc}, \bc(\tau+1)-\bc(\tau)\right\rangle + \bm{\chi}(t) + \bs(t)\right\|_2 \\
        = & \left\| -\eta \left \langle \frac{\partial \bs(\tau)}{\partial \ba}, \frac{\partial \mathcal{L}(\tau)}{\partial \ba}\right\rangle - \eta \left \langle \frac{\partial \bs(\tau)}{\partial \bc}, \frac{\partial \mathcal{L}(t)}{\partial \bc}\right\rangle + \bm{\chi}(t) + \bs(t)\right\|_2 \\
        \leq & \left\|\left( \bm{I} - \eta \bG(t)\right) \cdot \bs(t) \right\|_2 + \left\| \bm{\chi}(t) \right\|_2 \\
        \leq & \left(1- \frac{3\eta \sigma_{\min}}{4} \right) \left\| \bs(t) \right\|_2 + \frac{\eta \sigma_{\min}}{4} \left\| \bs(t)\right\|_2\\
        \leq & \left(1- \frac{\eta \sigma_{\min}}{2} \right) \left\| \bs(t) \right\|_2.
    \end{split}
\end{equation*}
Here, we further require that $\bm{I} - \eta \bG(t)$ is positive semi-definite. Since $\left\|\bG(t) - \bG^{\infty}\right\|_2 \leq \frac{\sigma_{\min}}{4}$, we have $\left\|\bG(t)\right\|_2 \leq \frac{5 \sigma_{\max}}{4}$, which implies $1- \eta \frac{5 \sigma_{\max}}{4} \geq 0$ and thus $\eta \leq \frac{4}{5 \sigma_{\max}} \lesssim \frac{1}{n_d}$.
Therefore, by induction, we have
\begin{equation*}
    \mathcal{L}(t) \leq \left(1- \frac{\eta \sigma_{\min}}{2} \right)^{t} \mathcal{L}(0).
\end{equation*}

\newpage
\section{Proofs for \Cref{sec_sgd}}
\label{appendix_proof_sgd}
\subsection{Proof for \Cref{lemma_local_sgd}}
\label{appendix_proof_lemma_local_sgd}
Following the same procedure and techniques as in \Cref{lemma_local}, we continue from \Cref{eq_appx_2} and obtain
\begin{equation*}
    \begin{split}
        & \left| G_{i,j}(\bm{a},\bm{c}) - G_{i,j}(\bm{a}(0),\bm{c}(0)) \right|\\
        \leq & \left|\sum_{q=1}^{m}\left\langle\frac{\partial G_{i,j}}{\partial \ba_q},\ba_{q} - \ba(0) \right\rangle + \left\langle \frac{\partial G_{i,j}}{\partial \bc_q},\bc_{q} - \bc(0)\right\rangle\right|\\
        \leq & \sum_{q=1}^{m} \left\|\frac{\partial G_{i,j}}{\partial \ba_q} \right\|_2 \cdot \left\| \ba_{q} - \ba(0)\right\|_2 + \left\|\frac{\partial G_{i,j}}{\partial \bc_q} \right\|_2 \cdot \left\| \bc_{q} - \bc(0)\right\|_2\\
        \lesssim & \sum_{q=1}^{m} \frac{n_d^{5/2} n^{3/2}}{mN} \left\|\bc_{q} \right\|_2^2 \cdot \left\| \ba_{q} - \ba(0)\right\|_2 + \frac{n_d^2 n}{mN} \left\|\bc_{q} \right\|_2 \cdot \left\| \bc_{q} - \bc(0)\right\|_2 \\
        \lesssim &  \frac{n_d^{5/2} n^{3/2}}{\sqrt{m} N} \widetilde{M}_c^2 \cdot R_a + \frac{n_d^2 n}{\sqrt{m} N} \widetilde{M}_c \cdot R_c.
    \end{split}
\end{equation*}
Let $\frac{n_d^{5/2} n^{3/2}}{\sqrt{m} N} \widetilde{M}_c^2 \cdot R_a \lesssim \frac{\sigma_{\min}}{16N}$ and $\frac{n_d^2 n}{\sqrt{m} N} \widetilde{M}_c \cdot R_c \lesssim \frac{\sigma_{\min}}{16N}$ such that 
\begin{equation*}
    \left| G_{i,j}(\bm{a},\bm{c}) - G_{i,j}(\bm{a}(0),\bm{c}(0)) \right| \leq \frac{\sigma_{\min}}{8 N},
\end{equation*}
which implies that 
\begin{equation*}
    \left\| \bG(\bm{a},\bm{c}) - \bG(\bm{a}(0),\bm{c}(0)) \right\| \leq N \cdot \frac{\sigma_{\min}}{8 N} \leq \frac{\sigma_{\min}}{8}.
\end{equation*}
Here, we require that
\begin{equation*}
    R_a \lesssim \frac{\sigma_{\min} \sqrt{m}}{n_d^{5/2} n^{3/2} \widetilde{M}_c^2},\quad\text{and}\quad R_c \lesssim \frac{\sigma_{\min} \sqrt{m}}{n_d^2 n \widetilde{M}_c}.
\end{equation*}
Let $\widetilde{M}_c \gtrsim \sqrt{n_d} + \sqrt{\log \frac{m}{\delta}}$ as in \Cref{eq_appx_3}, we have $R_a \lesssim R_c$. We alternatively require $\left\|\ba(t)-\ba(0) \right\|_2 \leq R$ and $\left\|\bc(t)-\bc(0) \right\|_2 \leq R$ for $R \lesssim \frac{\sigma_{\min} \sqrt{m}}{n_d^{5/2} n^{3/2} M_c^2}$.

\subsection{Proof for \Cref{lemma_expect_move}}
\label{appendix_proof_lemma_expect_move}
We first show a complementary result. Denote $\bm{f}$ as $[f_1, f_2, \cdots, f_N]$, the prediction of KAN for each sample. Then,
\begin{equation}
\label{eq_appx_15}
    \begin{split}
        \mathbb{E}_{\mathcal{I}}\left[\left\| \frac{\partial \Tilde{\mathcal{L}}(t)}{\partial \bm{f}}\right\|_2^2 \right] = & \mathbb{E}_{\mathcal{I}}\left[ \left\|\frac{2}{b} \sum_{i \in \mathcal{I}} (f_i - y_i) \bm{e}_{i}\right\|_2^2\right]\\
        = & \frac{4}{b} \mathbb{E}_{\mathcal{I}}\left[ \frac{1}{b} \sum_{i \in \mathcal{I}} (f_i - y_i)^2\right]\\
        = & \frac{4}{b} \cdot \frac{1}{N} \sum_{i=1}^{N} (f_i - y_i)^2\\
        = & \frac{4}{b} \mathcal{L}(t),
    \end{split}
\end{equation}
and
\begin{equation}
\label{eq_appx_16}
\begin{split}
    & \mathbb{E}\left[\left\| \ba(t+1) - \ba(t)\right\|_2^2 \cdot \bm{1}_{T > t}|\mathcal{F}_{t}\right] \\
    \leq & \eta^2 \mathbb{E}\left[\left.\left\| \frac{\partial \Tilde{\mathcal{L}}(t)}{\partial \ba}\right\|_2^2\right|\mathcal{F}_{t}, T>t\right]\\
    \leq & \eta^2 \left\| \left[\frac{\partial f_1}{\partial \ba},\cdots,\frac{\partial f_N}{\partial \ba}\right]\right\|_2^2 \cdot  \mathbb{E}\left[\left.\left\| \frac{\partial \Tilde{\mathcal{L}}(t)}{\partial \bm{f}}\right\|_2^2 \right| \mathcal{F}_{t}\right]\\
    \leq & \eta^2 \cdot  \left(\frac{5}{4}N\sigma_{\max}\right) \cdot \frac{4}{b} \mathcal{L}(t)\\
    \lesssim & \frac{N}{b} \eta^2 \sigma_{\max} \cdot \mathcal{L}(t) \lesssim \frac{N}{b} \eta^2 n_d \cdot \mathcal{L}(t).
\end{split}
\end{equation}
By a similar calculation, we have 
\begin{equation}
\label{eq_appx_17}
\begin{split}
    & \mathbb{E}\left[\left\| \bc(t+1) - \bc(t)\right\|_2^2\cdot \bm{1}_{T > t}|\mathcal{F}_{t}\right] \\
    \leq & \eta^2 \mathbb{E}\left[\left.\left\| \frac{\partial \Tilde{\mathcal{L}}(t)}{\partial \bc}\right\|_2^2\right|\mathcal{F}_{t}, T>t\right]\\
    \leq & \eta^2 \left\| \left[\frac{\partial f_1}{\partial \bc},\cdots,\frac{\partial f_N}{\partial \bc}\right]\right\|_2^2 \cdot \mathbb{E}\left[\left.\left\| \frac{\partial \Tilde{\mathcal{L}}(t)}{\partial \bm{f}}\right\|_2^2 \right| \mathcal{F}_{t}\right]\\
    \leq & \eta^2 \cdot  \left(\frac{5}{4}N\sigma_{\max}\right) \cdot \frac{4}{b} \mathcal{L}(t)\\
    \lesssim & \frac{N}{b} \eta^2 \sigma_{\max} \cdot \mathcal{L}(t) \lesssim \frac{N}{b} \eta^2 n_d \cdot \mathcal{L}(t).
\end{split}
\end{equation}

\subsection{Proof for \Cref{lemma_inball}}
\label{appendix_proof_lemma_inball}
\begin{equation}
\label{eq_appx_5}
    \begin{split}
        \left\|\bs(T-1)\right\|_2 = & \left\|\bs(0) + \left\langle \frac{\partial \bs}{\partial \ba}, \ba(T-1) - \ba(0)\right\rangle + \left\langle \frac{\partial \bs}{\partial \bc}, \bc(T-1) - \bc(0)\right\rangle\right\|_2\\
        \leq & \left\|\bs(0)\right\|_2 + \left\|\left[\begin{array}{ccc}
            \frac{\partial s_1}{\partial \ba} & \cdots & \frac{\partial s_N}{\partial \ba} \\
            \frac{\partial s_1}{\partial \bc}& \cdots & \frac{\partial s_N}{\partial \bc}
        \end{array} \right] \right\|_2 \cdot \left\| \left[\begin{array}{c}
            \ba(T-1)-\ba(0) \\
            \bc(T-1)-\bc(0)
        \end{array} \right]\right\|_2\\
        \leq & \left\|\bs(0)\right\|_2 + \sqrt{\sigma_{\max}} \cdot \sqrt{2}R\\
        \lesssim & \sqrt{n_d} \cdot R.
    \end{split}
\end{equation}

By the update scheme for $\ba$ and $\bc$, we have
\begin{equation*}
    \begin{split}
        & \left\|\ba(T) - \ba(T-1) \right\|_2 \\
        = & \eta \cdot \left\| \frac{\partial \Tilde{\mathcal{L}}(T-1)}{\partial \ba}\right\|_2\\
        \leq & \eta \cdot \left\| \left[\begin{array}{ccc}
            \frac{\partial f_1(T-1)}{\partial \ba} & \cdots  & \frac{\partial f_N(T-1)}{\partial \ba} 
        \end{array}\right]\right\|_2 \cdot \left\|\frac{\partial \Tilde{\mathcal{L}}(T-1)}{\partial \bm{f}} \right\|_2\\
        \leq & \eta \cdot \frac{N}{b} \sqrt{\sigma_{\max}} \sqrt{\mathcal{L}(T-1)} \\
        \lesssim & \eta \cdot \frac{N}{b} n_d R.
    \end{split}
\end{equation*}
We let $\eta \cdot \frac{N}{b} n_d R \lesssim \frac{1}{2}R$ such that $\left\|\ba(T) - \ba(T-1) \right\|_2 \leq \frac{1}{2}R$ and $\left\|\ba(T) - \ba(0) \right\|_2 \leq R$, where we require that
\begin{equation*}
    \eta \lesssim \frac{b}{N n_d}. 
\end{equation*}
The similar result also holds for $\left\|\bc(T) - \bc(0) \right\|_2 \leq R$. 

\begin{equation*}
    \begin{split}
        & \left\|\bc_{q}(T) - \bc_{q}(T-1) \right\|_2 \\
        = & \eta \cdot \left\| \frac{\partial \Tilde{\mathcal{L}}(T-1)}{\partial \bc_{q}}\right\|_2\\
        \lesssim & \eta \cdot \frac{N \sqrt{n_d}}{\sqrt{m}} \sqrt{\mathcal{L}(T-1)}\\
        \lesssim & \eta \cdot \frac{N n_d}{b \sqrt{m}} R \\
        \lesssim & \eta \cdot \frac{N \sigma_{\min}}{b n_d^{3/2} n^{3/2} \widetilde{M}_c^2}.
    \end{split}
\end{equation*}
Let $\eta \cdot \frac{N \sigma_{\min}}{b n_d^{3/2} n^{3/2} \widetilde{M}_c^2} \lesssim \frac{1}{2} \widetilde{M}_c$ such that $\left\|\bc_{q}(T) - \bc_{q}(T-1) \right\|_2 \leq \frac{1}{2} \widetilde{M}_c$ and $\left\|\bc_{q}(T)\right\|_2 \leq \widetilde{M}_c$, under the condition that 
\begin{equation*}
    \eta \lesssim \frac{b n_d^{3/2} n^{3/2} \widetilde{M}_c^3}{N  \sigma_{\min}}.
\end{equation*}
Therefore, we require
\begin{equation*}
    \eta \lesssim \frac{b}{N n_d}. 
\end{equation*}

\subsection{Proof for \Cref{lemma_error_sgd}}
\label{appendix_proof_lemma_error_sgd}
Under the condition of $T>t$, we continue from \Cref{eq_appx_4} that
\begin{equation*}
    \begin{split}
    & \left|\chi_i(t)\right|\\
    \lesssim &  \sum_{q=1}^{m} \sqrt{\frac{n_d n}{m N}} \left\|\bc_{q} \right\|_2 \cdot \left\| \ba_q(t+1) - \ba_q(t)\right\|_2^2 + \sqrt{\frac{1}{mN}} \cdot \left( \left\| \ba_q(t+1) - \ba_q(t)\right\|_2^2 + \left\| \bc_q(t+1) - \bc_q(t)\right\|_2^2\right),\\
    \lesssim & \sqrt{\frac{n_d n}{m N}} \widetilde{M}_c \cdot \left( \left\| \ba(t+1) - \ba(t)\right\|_2^2 + \left\| \bc(t+1) - \bc(t)\right\|_2^2 \right),
    \end{split}
\end{equation*}
then
\begin{equation*}
    \left\|\bm{\chi}(t) \right\|_2 \lesssim \sqrt{\frac{n_d n}{m}} \widetilde{M}_c \cdot \left( \left\| \ba(t+1) - \ba(t)\right\|_2^2 + \left\| \bc(t+1) - \bc(t)\right\|_2^2 \right).
\end{equation*}
Therefore, we obtain
\begin{equation*}
        \mathbb{E}\left[ \left\|\bm{\chi}(t) \right\|_2 \cdot \bm{1}_{T > t}|\mathcal{F}_{t}\right]
        \lesssim  \sqrt{\frac{n_d^{3} n}{m}} \widetilde{M}_c \cdot \frac{N}{b} \eta^2  \mathcal{L}(t),
\end{equation*}
and
\begin{equation*}
        \mathbb{E}\left[ \left\|\bm{\chi}(t) \right\|_2 \cdot \left\| \bs(t)\right\|_2\cdot \bm{1}_{T > t}|\mathcal{F}_{t}\right]
        \lesssim \frac{n_d^{2} \sqrt{n}}{\sqrt{m}}  \widetilde{M}_c \cdot \frac{N}{b} \eta^2   R \cdot \mathcal{L}(t),
\end{equation*}
where the last inequality is due to \Cref{eq_appx_5}. Let 
\begin{equation*}
    \frac{n_d^{2} \sqrt{n}}{\sqrt{m}}  \widetilde{M}_c \cdot \frac{N}{b} \eta^2   R \cdot \mathcal{L}(t) \lesssim  \frac{\eta \sigma_{\min}}{8} \cdot \mathcal{L}(t),
\end{equation*}
such that $\mathbb{E}\left[ \left\|\bm{\chi}(t) \right\|_2 \cdot \left\| \bs(t)\right\|_2\right] \leq \frac{\eta \sigma_{\min}}{8} \cdot \mathcal{L}(t)$, where we require
\begin{equation*}
    \eta \lesssim \frac{b}{N} \cdot \frac{n \widetilde{M}_c}{\sqrt{n_d}}.
\end{equation*}

Furthermore, when $T>t$, we have
\begin{equation*}
    \begin{split}
        & \left\|\bs(t+1) - \bs(t) \right\|_2 \\
        = & \left\|\left[\begin{array}{ccc}
            \frac{\partial s_1}{\partial \ba} & \cdots & \frac{\partial s_N}{\partial \ba} \\
            \frac{\partial s_1}{\partial \bc}& \cdots & \frac{\partial s_N}{\partial \bc}
        \end{array} \right] \right\|_2 \cdot \left\| \left[\begin{array}{c}
            \ba(t+1)-\ba(t) \\
            \bc(t+1)-\bc(t)
        \end{array} \right]\right\|_2\\
        \leq & \sqrt{\sigma_{\max}}  \cdot \left\| \left[\begin{array}{c}
            \ba(t+1)-\ba(t) \\
            \bc(t+1)-\bc(t)
        \end{array} \right]\right\|_2,
    \end{split}
\end{equation*}
and
\begin{equation*}
    \begin{split}
        & \mathbb{E}\left[\left.\left\|\bs(t+1) - \bs(t) \right\|_2^2 \cdot \bm{1}_{T > t}\right|\mathcal{F}_{t}\right]\\
        \leq & \sigma_{\max} \cdot \mathbb{E}\left[\left.\left(\left\|\ba(t+1)-\ba(t)\right\|_2^2 + \left\|\bc(t+1)-\bc(t)\right\|_2^2\right) \cdot \bm{1}_{T > t}\right|\mathcal{F}_{t}\right]\\
        \lesssim & \frac{N}{b} \eta^2 n_d^{2} \cdot \mathcal{L}(t).
    \end{split}
\end{equation*}
We let $\frac{N}{b} \eta^2 n_d^{2} \lesssim \frac{\eta \sigma_{\min}}{8}$ such that $\mathbb{E}\left[\left.\left\|\bs(t+1) - \bs(t) \right\|_2^2 \cdot \bm{1}_{T > t}\right|\mathcal{F}_{t}\right] \leq  \frac{\eta \sigma_{\min}}{8}$, where we require
\begin{equation*}
    \eta \lesssim \frac{b}{N} \cdot \frac{\sigma_{\min}}{n_d^{2}}.
\end{equation*}

\subsection{Proof for \Cref{lemma_convg_sgd}}
\label{appendix_proof_lemma_convg_sgd}
\begin{equation*}
    \begin{split}
        \mathcal{L}(t+1) = & \left\|\bs(t+1) \right\|_2^2 \\
        = & \left\|\bs(t+1) -\bs(t) + \bs(t)\right\|_2^2 \\
        = & 2 \left \langle \frac{\partial \bs(t)}{\partial \ba}\left(\ba(t+1)-\ba(t)\right) +\frac{\partial \bs(t)}{\partial \bc}\left(\bc(t+1)-\bc(t), \bs(t) \right)\right\rangle \\
        & + 2\left\langle \bm{\chi}(t), \bs(t) \right\rangle + \left\|\bs(t+1)-\bs(t) \right\|_2^2 + \left\|\bs(t) \right\|_2^2,
    \end{split}
\end{equation*}
then
\begin{equation*}
    \begin{split}
        \mathbb{E}\left[
        \left.\mathcal{L}(t+1)\cdot \bm{1}_{T > t}\right|\mathcal{F}_{t}\right] \leq & - 2\eta \cdot \bs(t) \bG(t) \bs(t) + 2 \mathbb{E}\left[\left.\left\| \bm{\chi}(t)\right\|_2 \cdot \left\| \bs(t)\right\|_2 \cdot \bm{1}_{T > t}\right|\mathcal{F}_{t}\right] \\
        & + \mathbb{E}\left[\left.\left\|\bs(t+1) - \bs(t) \right\|_2^2 \cdot \bm{1}_{T > t}\right|\mathcal{F}_{t}\right] + \left\| \bs(t) \right\|_2^2\\
        \leq & \left(1 - \eta \sigma_{\min}\right) \cdot \mathcal{L}(t).
    \end{split}
\end{equation*}
Therefore, by induction, we have
\begin{equation*}
    \begin{split}
        & \mathbb{E}\left[\mathcal{L}(t+1) \cdot \bm{1}_{T > t}\right]\\
        = & \mathbb{E}\left[ \mathbb{E}\left[\mathcal{L}(t+1) \cdot \bm{1}_{T \geq t}|\mathcal{F}_{t}\right]\right]\\
        \leq & \mathbb{E}\left[\left(1-\eta \sigma_{\min}\right) \mathcal{L}(t) \cdot \bm{1}_{T \geq t}\right] \\
        \leq & \left(1-\eta \sigma_{\min}\right) \mathbb{E}\left[ \mathcal{L}(t) \cdot \bm{1}_{T \geq t-1}\right] \\
        \leq & \cdots \\
        \leq &  \left(1 - \eta \sigma_{\min}\right)^{t} \cdot \mathcal{L}(0).
    \end{split}
\end{equation*}

\subsection{Proof for \Cref{lemma_event_never_leave_ball}}
\label{appendix_proof_lemma_event_never_leave_ball}
\begin{equation*}
    \begin{split}
        & \mathbb{E}\left[\sum_{t=0}^{T-1} \left\|\ba(t+1)-\ba(t) \right\|_2\right] \\
        = & \mathbb{E}\left[\sum_{t=0}^{T-1} \left. \left\|\ba(t+1)-\ba(t) \right\|_2 \cdot \bm{1}_{T > t}\right|\mathcal{F}_{t}\right] \\
        = & \mathbb{E}\left[\sum_{t=0}^{T-1} \mathbb{E}\left[\left.\left\|\ba(t+1)-\ba(t) \right\|_2 \cdot \bm{1}_{T > t}\right|\mathcal{F}_{t}\right]\right] \\
        \leq & \mathbb{E}\left[\sum_{t=0}^{T-1} \sqrt{\mathbb{E}\left[\left.\left\|\ba(t+1)-\ba(t) \right\|_2^2 \cdot \bm{1}_{T > t}\right|\mathcal{F}_{t}\right]}\right]\\
        \lesssim & \eta \cdot \sqrt{\frac{N n_d}{b}} \cdot \mathbb{E}\left[\sum_{t=0}^{T-1} \sqrt{\mathcal{L}(t) \cdot \bm{1}_{T > t}}\right]\\
        \leq & \eta \cdot \sqrt{\frac{N n_d}{b}} \cdot \mathbb{E}\left[\sum_{t=0}^{T-1} \mathbb{E}\left[\sqrt{\mathcal{L}(t) \cdot \bm{1}_{T > t-1}} \right]\right]\\
        \lesssim & \frac{\sqrt{n_d}}{\sigma_{\min}} \sqrt{\frac{N}{b}} \cdot \sqrt{\mathcal{L}(0)}.
    \end{split}
\end{equation*}

Similarly, we have
\begin{equation*}
    \begin{split}
        & \mathbb{E}\left[\sum_{t=0}^{T-1} \left\|\bc(t+1)-\bc(t) \right\|_2\right] \\
        \lesssim & \frac{\sqrt{n_d}}{\sigma_{\min}} \sqrt{\frac{N}{b}} \cdot \sqrt{\mathcal{L}(0)}.
    \end{split}
\end{equation*}
Then,
\begin{equation*}
\begin{split}
    & \mathbb{P}\left(\left\| \ba(T) -\ba(0)\right\|_2 > \frac{1}{2}R\right)\\
    \leq & \mathbb{P}\left(\sum_{t=0}^{T-1} \left\|\ba(t+1)-\ba(t) \right\|_2 > R\right) \\
    \leq & \frac{\mathbb{E}\left[\sum_{t=0}^{T-1} \left\|\ba(t+1)-\ba(t) \right\|_2\right]}{R} \\
    \lesssim & \frac{\frac{\sqrt{n_d}}{\sigma_{\min}} \sqrt{\frac{N}{b}} \sqrt{\mathcal{L}(0)}}{\frac{\sigma_{\min} \sqrt{m}}{n_d^{5/2} n^{3/2} \widetilde{M}_c^2}}.
\end{split}
\end{equation*}
We let the last term in the above inequality be less than $\frac{1}{3}\Tilde{\delta}$, such that 
\begin{equation*}
    \mathbb{P}\left(\left\| \ba(T) -\ba(0)\right\|_2 > \frac{1}{2}R\right) \leq \frac{1}{3} \Tilde{\delta},
\end{equation*}
under the condition that
\begin{equation}
\label{eq_appx_6}
    m \gtrsim \frac{N n_d^6 n^3 \widetilde{M}_c^4}{b \sigma_{\min}^{4} \Tilde{\delta}^2} \cdot \mathcal{L}(0).
\end{equation}
Similar results also hold for
\begin{equation*}
    \mathbb{P}\left(\left\| \bc(T) -\bc(0)\right\|_2 > \frac{1}{2}R\right) \leq \frac{1}{3} \Tilde{\delta}.
\end{equation*}

Moreover, for all $q \in [m]$, 
\begin{equation*}
   \begin{split}
        & \left\| \bc_{q}(T)\right\|_2 \\
        \leq & \left\| \bc_{q}(T) - \bc_{q}(0)\right\|_2 + \left\|\bc_{q}(0)\right\|_2\\
        \leq & \left\| \bc(T) - \bc(0)\right\|_2 + M_c.
   \end{split} 
\end{equation*}
Therefore, we have
\begin{equation*}
    \begin{split}
        & \mathbb{P}\left(\exists q \in [m], \left\| \bc_{q}(T)\right\|_2 > \frac{1}{2} \widetilde{M}_c \right)\\
        \leq & \mathbb{P}\left(\left\| \bc(T) - \bc(0)\right\|_2 +  M_c > \frac{1}{2} \widetilde{M}_c \right)\\
        \leq &  \mathbb{P}\left(\sum_{t=0}^{T-1} \left\|\bc(t+1)-\bc(0) \right\|_2 + M_c> \frac{1}{2} \widetilde{M}_c \right)\\
        \leq & \frac{\mathbb{E}\left[\sum_{t=0}^{T-1} \left\|\bc(t+1)-\bc(0) \right\|_2\right] + M_c}{\frac{1}{2} \widetilde{M}_c}.
    \end{split}
\end{equation*}
Let the last term of the above inequality be less than $\frac{1}{3}\Tilde{\delta}$, we have
\begin{equation*}
    \widetilde{M}_c \simeq \frac{\frac{\sqrt{n_d}}{\sigma_{\min}} \sqrt{\frac{N}{b}} \cdot \sqrt{\mathcal{L}(0)} + M_c}{\Tilde{\delta}}.
\end{equation*}
Note that $\left\| \ba(T) -\ba(0)\right\|_2 \leq \frac{1}{2}R$, $\left\| \bc(T) -\bc(0)\right\|_2 \leq \frac{1}{2}R$, and $\left\| \bc_{q}(T)\right\|_2 \leq \frac{1}{2} \widetilde{M}_c $ for all $q \in [m]$ implies that $T = \infty$, where the sequence generated by SGD never leaves the neighborhood. Plugging the above $\widetilde{M}_c$ into \Cref{eq_appx_6}, we get the desired results. 

\subsection{Proof for \Cref{theorem_sgd}}
\label{appendix_proof_theorem_sgd}
Summarizing Lemmas \ref{lemma_local_sgd}-\ref{lemma_event_never_leave_ball}, we conclude that
\begin{equation*}
    \begin{split}
        & \mathbb{E}\left[\mathcal{L}(t+1) \cdot \bm{1}_{\mathcal{E}}\right] \\
        \leq & \mathbb{E}\left[\mathcal{L}(t+1) \cdot \bm{1}_{T > t}\right]\\
        = & \mathbb{E}\left[ \mathbb{E}\left[\mathcal{L}(t+1) \cdot \bm{1}_{T > t}|\mathcal{F}_{t}\right]\right]\\
        \leq & \mathbb{E}\left[\left(1-\eta \sigma_{\min}\right) \mathcal{L}(t) \cdot \bm{1}_{T > t}\right] \\
        \leq & \left(1-\eta \sigma_{\min}\right) \mathbb{E}\left[ \mathcal{L}(t) \cdot \bm{1}_{T > t-1}\right] \\
        \leq & \cdots \\
        \leq &  \left(1 - \eta \sigma_{\min}\right)^{t} \cdot \mathcal{L}(0),
    \end{split}
\end{equation*}
under the conditions
\begin{equation*}
    m \gtrsim \frac{n_d^4 n^2}{\sigma_{\min}^2} \log \left(\frac{N}{\delta}\right) \cdot \left(n_d + \log \frac{m}{\delta}\right)
\end{equation*}
from \Cref{lemma_init_gram},
and
\begin{equation*}
    \begin{split}
        m \gtrsim & \frac{N^3  n_d^{11} n^3}{b^3 \sigma_{\min}^{8} \Tilde{\delta}^{6}} \cdot \left(\log \frac{N}{\delta}\right)^{3}\\
        & + \frac{N n_d^7 n^3}{b \sigma_{\min}^4 \Tilde{\delta}^6} \cdot \left(n_d^2 + \left(\log \frac{m}{\delta}\right)^{2} \right)\cdot \left(\log \frac{N}{\delta}\right)
    \end{split}
\end{equation*}
from \Cref{lemma_event_never_leave_ball}.

\newpage
\section{Proof for \Cref{sec_pikan}}
\label{appendix_proof_pikan}
Here, we introduce additional notations for physics-informed KANs. For example, the physics-informed loss function is defined as
\begin{equation*}
\begin{split}
    \mathcal{L}^{\text{PDE}}(\ba,\bc)  = & \frac{1}{N_1}\sum_{i=1}^{N_1} \left( \mathcal{D}[f(\ba,\bc;\bx_{i})] - v_i \right)^2 + \frac{1}{N_2}\sum_{i=1}^{N_2} \left( f(\ba,\bc;\Bar{\bx}_{i}) - u_i \right)^2\\
    := & \sum_{i=1}^{N_1} s_i^2 + \sum_{i=1}^{N_2} \Bar{s}_i^2,
\end{split}
\end{equation*}
which follows the similar notation as the regression loss. The matrix $\bD$ and the corresponding Gram matrix are defined as
\begin{equation*}
    \bD = \left[\begin{array}{cccccc}
        \frac{\partial s_1}{\partial \ba} & \cdots & \frac{\partial s_{N_1}}{\partial \ba} & \frac{\partial \Bar{s}_1}{\partial \ba} & \cdots & \frac{\partial \Bar{s}_{N_2}}{\partial \ba}\\
        \frac{\partial s_1}{\partial \bc} & \cdots & \frac{\partial s_{N_1}}{\partial \bc} & \frac{\partial \Bar{s}_1}{\partial \bc} & \cdots & \frac{\partial \Bar{s}_{N_2}}{\partial \bc}\\
    \end{array} \right],
\end{equation*}
and
\begin{equation*}
    \bG = \bD^{\top} \bD.
\end{equation*}

\subsection{Proof for \Cref{theorem_gd_pde}}
Since the proof structure is similar and detailed computation is complicated, we provide key components of the proof here, which are sufficient to derive and verify the final results.
In \Cref{appendix_proof_lemma_init_gram}, \Cref{eq_appx_7} is modified as
$\left| X_q \right| \leq \frac{1}{N}n_d^2 n_3 \left\| \bc_{q} \right\|_2^2 \cdot \left\| \ba_{q} \right\|_2^4$. 
Therefore, by the Hoeffding's inequality, we have 
\begin{equation*}
    \begin{split}
        & \mathbb{P}\left(\left|S_{i,j}(0) - \mathbb{E}\left[S_{i,j} \right]\right| > \epsilon \right) \\
        = & \mathbb{P}\left(\left|\frac{1}{m} \sum_{q=1}^{m} X_{q} - \mathbb{E}\left[X_{q} \right]\right| > \epsilon \right)\\
        \leq & 2 \exp\left(-\frac{m \epsilon^2 N_1^2}{2 n_d^4 n^6 M_a^{8} M_c^4}\right),
    \end{split}
\end{equation*}
as a counterpart of \Cref{eq_appx_8}.
Let $\epsilon = \frac{\sigma_{\min}}{8N_1}$ and set the last term in the above inequality to be less than $\frac{\delta}{4N_1^2}$, then with a probability of at least $1-\frac{\delta}{4N_1^2}$, we have
\begin{equation*}
    \left|S_{i,j}(0) - \mathbb{E}\left[S_{i,j} \right]\right| \leq \frac{\sigma_{\min}}{16N},
\end{equation*}
under the condition that
\begin{equation*}
    m \gtrsim \frac{n_d^4 n^6}{\sigma_{\min}^2} \log \left(\frac{N}{\delta}\right) \cdot \left(n_d^2 + \left(\log \frac{m}{\delta}\right)^2\right) \cdot \left(n_d^4 n^4 + \left(\log \frac{m}{\delta}\right)^4\right).
\end{equation*}
In summary, \Cref{lemma_init_gram} holds for physics-informed KANs under the above condition on the hidden dimension $m$.

Next, we proceed to prove \Cref{lemma_local} for physics-informed KANs, where we additionally require that
\begin{equation*}
    \left\|\ba_{q} \right\|_2 \leq M_a,
\end{equation*}
for all $q \in [m]$. Following \Cref{appendix_proof_lemma_local}, Equations \ref{eq_appx_9}-\ref{eq_appx_12} are modified as
\begin{equation*}
    \left|\frac{\partial S_{i,j}}{\partial a_{p,q,k}} \right| \leq \frac{n_d^2 n^3}{m N} \cdot \left\| \bc_{q} \right\|_2^2 \cdot \left\| \ba_{q} \right\|_2^4,
\end{equation*}
\begin{equation*}
    \left\|\frac{\partial S_{i,j}}{\partial \ba_{q}} \right\|_2 \leq \frac{n_d^{5/2} n^{7/2}}{m N} \cdot \left\| \bc_{q} \right\|_2^2 \cdot \left\| \ba_{q} \right\|_2^4,
\end{equation*}
\begin{equation*}
    \left| \frac{\partial S_{i,j}}{\partial c_{q,k}}\right| \lesssim \frac{n_d^{3/2} n^{3}}{mN} \left\|\bc_{q} \right\|_2 \cdot \left\| \ba_{q} \right\|_2^4,
\end{equation*}
and 
\begin{equation*}
    \left\| \frac{\partial S_{i,j}}{\partial \bc_{q}}\right\|_2 \leq \frac{n_d^2 n^{3}}{mN} \left\|\bc_{q} \right\|_2 \cdot \left\| \ba_{q} \right\|_2^4,
\end{equation*}
respectively. Then \Cref{eq_appx_2} corresponds to
\begin{equation}
\label{eq_appx_13}
    \begin{split}
        & \left| G_{i,j}(\bm{a},\bm{c}) - G_{i,j}(\bm{a}(0),\bm{c}(0)) \right|\\
        \lesssim & \frac{n_d^{5/2} n^{7/2}}{\sqrt{m}N} M_a^4 M_c^2 \cdot R_a + \frac{n_d^2 n^3}{\sqrt{m} N} M_a^4 M_c  \cdot R_c.
    \end{split}
\end{equation}
Let 
\begin{equation*}
    \frac{n_d^{5/2} n^{7/2}}{\sqrt{m}N} M_a^4 M_c^2 \cdot R_a + \frac{n_d^2 n^3}{\sqrt{m} N} M_a^4 M_c \cdot R_c \lesssim \frac{\sigma_{\min}}{8N},
\end{equation*}
such that $\left| G_{i,j}(\bm{a},\bm{c}) - G_{i,j}(\bm{a}(0),\bm{c}(0)) \right| \leq \frac{\sigma_{\min}}{8N_1}$, then we equivalently require that 
\begin{equation*}
    R_a \lesssim \frac{\sigma_{\min} \sqrt{m}}{n_d^{5/2} n^{7/2} M_a^4 M_c^2},\quad\text{and}\quad R_c \lesssim \frac{\sigma_{\min}\sqrt{m}}{n_d^2 n^3 M_a^4 M_c}.
\end{equation*}
Therefore, we have
\begin{equation*}
    \left\|\bG(\ba,\bc) - \bG(\ba(0),\bc(0)) \right\|_2 \leq N \cdot \frac{\sigma_{\min}}{8N} \leq \frac{\sigma_{\min}}{8}.
\end{equation*}
In summary, \Cref{lemma_local} holds for physics-informed KANs with the aforementioned $R_a$, $R_c$, $M_a$ and $M_c$.

The initialized loss $\mathcal{L}^{\text{PDE}}(0)$ is bounded, using the same technique as \Cref{lemma_init_loss}. Specifically, we have
\begin{equation*}
    \mathcal{L}^{\text{PDE}}(0) \lesssim n_d^2 n^4 \cdot \left(\log \frac{N_1 + N_2}{\delta}\right).
\end{equation*}
Note that 
\begin{equation*}
    G_{i,j}^{\infty} = \mathbb{E}\left[G_{i,j}\right] \lesssim \mathbb{E}\left[\frac{n^2}{m N} \sum_{q=1}^{m} \left(\left\| \bc_q\right\|_2^2 \cdot \left\| \ba_q\right\|_2^4 \right)\right] \lesssim \frac{n_d^3 n^4}{N},
\end{equation*}
which implies that $\sigma_{\max} \lesssim n_d^3 n^4$.

Now we are ready to extend \Cref{lemma_next_step_inball} to physics-informed KANs, a key lemma that guarantees $\ba$ and $\bc$ remain within the neighborhoods of their initialization.
\begin{equation*}
    \begin{split}
        \left\|\ba(t+1) - \ba(0) \right\|_2 \leq &  \sum_{\tau=0}^{t}\left\|\ba(\tau+1) - \ba(\tau) \right\|_2 \\
        \lesssim & \frac{\sqrt{\sigma_{\max}} }{\sigma_{\min}} \cdot \sqrt{\mathcal{L}^{\text{PDE}}(0)} \lesssim \frac{n_d^{3/2} n^2}{\sigma_{\min}} \cdot \sqrt{\mathcal{L}^{\text{PDE}}(0)},
    \end{split}
\end{equation*}
and
\begin{equation*}
    \left\|\bc(t+1) - \bc(0) \right\|_2 \lesssim \frac{n_d^{3/2} n^2}{\sigma_{\min}} \cdot \sqrt{\mathcal{L}^{\text{PDE}}(0)},
\end{equation*}
where $\sqrt{\mathcal{L}^{\text{PDE}}(0)} = \left\| \left[\begin{array}{c}
    \bs(0) \\
    \Bar{\bs}(0)
\end{array} \right]\right\|_2$.
We have
\begin{equation*}
    \begin{split}
        \left\|\ba_q(t+1) - \ba_q(0) \right\|_2 \leq & \eta \sum_{\tau=0}^{t}\left\|\ba_q(\tau+1) - \ba_q(\tau) \right\|_2 \\
        \leq & \eta \sum_{\tau=0}^{t} \left\| \frac{\partial \mathcal{L}(\tau)}{\partial \ba_q}\right\|_2 \\
        \leq & \eta \sum_{\tau=0}^{t} \left( \sum_{i=1}^{N_1}\left|s_i(\tau)\right| \cdot \left\| \frac{\partial s_i(\tau)}{\partial \ba_q}\right\|_2 + \sum_{i=1}^{N_2}\left|\Bar{s}_i(\tau)\right| \cdot \left\| \frac{\partial \Bar{s}_i(\tau)}{\partial \ba_q}\right\|_2\right)\\
        \lesssim & \eta \frac{n_d n^{3/2} M_a^2 M_c}{\sqrt{m}} \sum_{\tau=0}^{t} \left\| \bs(\tau)\right\|_2 + \left\| \Bar{\bs}(\tau)\right\|_2\\
        \lesssim & \eta \frac{n_d n^{3/2} M_a^2 M_c}{\sqrt{m}}\sum_{\tau=0}^{t} \left(1- \eta \cdot \frac{\sigma_{\min}}{2} \right)^{\tau} \cdot \sqrt{\mathcal{L}^{\text{PDE}}(0)}\\
        \lesssim & \frac{n_d n^{3/2} M_a^2 M_c}{\sqrt{m} \sigma_{\min}} \cdot \sqrt{\mathcal{L}^{\text{PDE}}(0)},
    \end{split}
\end{equation*}
and
\begin{equation*}
    \begin{split}
        \left\|\bc_q(t+1) - \bc_q(0) \right\|_2 \leq & \eta \sum_{\tau=0}^{t}\left\|\bc_q(\tau+1) - \bc_q(\tau) \right\|_2 \\
        \leq & \eta \sum_{\tau=0}^{t} \left\| \frac{\partial \mathcal{L}(\tau)}{\partial \bc_q}\right\|_2 \\
        \leq & \eta \sum_{\tau=0}^{t} \left( \sum_{i=1}^{N_1}\left|s_i(\tau)\right| \cdot \left\| \frac{\partial s_i(\tau)}{\partial \bc_q}\right\|_2 + \sum_{i=1}^{N_2}\left|\bar{s}_i(\tau)\right| \cdot \left\| \frac{\partial \Bar{s}_i(\tau)}{\partial \bc_q}\right\|_2\right)\\
        \lesssim & \eta \frac{\sqrt{n_d}n M_a^2}{\sqrt{m}} \sum_{\tau=0}^{t} \left\| \bs(\tau)\right\|_2 + \left\| \bar{\bs}(\tau)\right\|_2\\
        \lesssim & \eta \frac{\sqrt{n_d}n M_a^2}{\sqrt{m}}\sum_{\tau=0}^{t} \left(1- \eta \cdot \frac{\sigma_{\min}}{2} \right)^{\tau} \cdot \sqrt{\mathcal{L}^{\text{PDE}}(0)}\\
        \lesssim & \frac{\sqrt{n_d} n M_a^2}{\sqrt{m} \sigma_{\min}} \cdot \sqrt{\mathcal{L}^{\text{PDE}}(0)},
    \end{split}
\end{equation*}

Let
\begin{equation*}
    \frac{n_d^{3/2} n^2}{\sigma_{\min}} \cdot \sqrt{\mathcal{L}^{\text{PDE}}(0)} \lesssim R_a,~\frac{n_d^{3/2} n^2}{\sigma_{\min}} \cdot \sqrt{\mathcal{L}^{\text{PDE}}(0)} \lesssim R_c,
\end{equation*}
and
\begin{equation*}
    \frac{n_d n^{3/2} M_a^2 M_c}{\sqrt{m} \sigma_{\min}} \cdot \sqrt{\mathcal{L}^{\text{PDE}}(0)} \lesssim M_a,~\frac{\sqrt{n_d} n M_a^2}{\sqrt{m} \sigma_{\min}} \cdot \sqrt{\mathcal{L}^{\text{PDE}}(0)} \lesssim M_c,
\end{equation*}
such that
\begin{equation*}
    \left\|\ba(t+1) - \ba(0) \right\|_2 \leq R_a,~\left\|\bc(t+1) - \bc(0) \right\|_2 \leq R_c,
\end{equation*}
and
\begin{equation*}
    \left\|\ba_{q}(t+1)\right\|_2 \leq M_a,~\left\|\bc_{q}(t+1)\right\|_2 \leq M_c,
\end{equation*}
we require
\begin{equation*}
    m \gtrsim \frac{n_d^{10} n^{15}}{\sigma_{\min}^{4}} \left(n_d^{6} n^{4} + \left(\log \frac{m}{\delta} \right)^6 \right) \cdot \left(\log \frac{N_1 + N_2}{\delta}\right),
\end{equation*}
where we plug the results of $\mathcal{L}^{\text{PDE}}(0)$ into the above inequality.

\subsection{Proof for \Cref{theorem_sgd_pde}}
\Cref{eq_appx_13} shows that \Cref{lemma_local_sgd} holds for physics-informed KANs with
\begin{equation*}
    R \lesssim \frac{\sigma_{\min} \sqrt{m}}{n_d^{5/2} n^{7/2} \widetilde{M}_a^4 \widetilde{M}_c^2},
\end{equation*}
where
\begin{equation*}
    \left\|\ba_q \right\|_2 \leq \widetilde{M}_a,\quad\text{and}\quad  \left\|\bc_q \right\|_2 \leq \widetilde{M}_c,
\end{equation*}
for all $q \in [m]$.

Analogous to \Cref{eq_appx_15}, we have
\begin{equation}
    \begin{split}
        \mathbb{E}_{\mathcal{I}}\left[\left\| \frac{\partial \Tilde{\mathcal{L}}(t)}{\partial \bm{s}}\right\|_2^2 \right] = & \mathbb{E}_{\mathcal{I}}\left[ \left\|\frac{2\sqrt{N_1}}{b_1} \sum_{i \in \mathcal{I}} s_i\bm{e}_{i} + \frac{2\sqrt{N_2}}{b_2}\sum_{i \in \mathcal{I}^{\prime} } \Bar{s}_i^2\bm{e}_i\right\|_2^2\right]\\
        = & \frac{4N_1}{b_1} \mathbb{E}_{\mathcal{I}}\left[ \frac{1}{b_1} \sum_{i \in \mathcal{I}} s_i^2\right] + \frac{4N_2}{b_2}\mathbb{E}_{\mathcal{I}^{\prime}}\left[ \frac{1}{b_2} \sum_{i \in \mathcal{I}^{\prime}} \Bar{s}_i^2\right] \\
        = & \frac{4N_1}{b_1} \cdot \frac{1}{N_1} \sum_{i=1}^{N} s_i^2 + \frac{4N_2}{b_2} \cdot \frac{1}{N_2} \sum_{i=1}^{N_2} \Bar{s}_i^2 \\
        = & \frac{4N_1}{b_1} \mathcal{L}(t),
    \end{split}
\end{equation}
where we require $\frac{b_1}{N_1} = \frac{b_2}{N_2}$. If $\frac{b_1}{N_1} \neq \frac{b_2}{N_2}$, the estimation for the expectation becomes much more complicated and less beneficial for our proof.
Equations \ref{eq_appx_16} and \ref{eq_appx_17} correspond to
\begin{equation*}
\begin{split}
    & \mathbb{E}\left[\left\| \ba(t+1) - \ba(t)\right\|_2^2 \cdot \bm{1}_{T > t}|\mathcal{F}_{t}\right] \\
    \leq & \eta^2 \mathbb{E}\left[\left.\left\| \frac{\partial \Tilde{\mathcal{L}}(t)}{\partial \ba}\right\|_2^2\right|\mathcal{F}_{t}, T>t\right]\\
    \lesssim & \frac{N}{b} \eta^2 \sigma_{\max} \cdot \mathcal{L}(t) \lesssim \frac{N}{b} \eta^2 n_d^{3} n^4 \cdot \mathcal{L}(t),
\end{split}
\end{equation*}
and
\begin{equation*}
\begin{split}
    & \mathbb{E}\left[\left\| \bc(t+1) - \bc(t)\right\|_2^2 \cdot \bm{1}_{T > t}|\mathcal{F}_{t}\right] \\
    \leq & \eta^2 \mathbb{E}\left[\left.\left\| \frac{\partial \Tilde{\mathcal{L}}(t)}{\partial \bc}\right\|_2^2\right|\mathcal{F}_{t}, T>t\right]\\
    \lesssim & \frac{N}{b} \eta^2 \sigma_{\max} \cdot \mathcal{L}(t) \lesssim \frac{N}{b} \eta^2 n_d^{3} n^4 \cdot \mathcal{L}(t),
\end{split}
\end{equation*}
respectively.

Now, we are ready to extend \Cref{lemma_inball} (with proof in \Cref{appendix_proof_lemma_inball}) for physics-informed KANs:
\begin{equation}
        \left\|\bs(T-1)\right\|_2 
        \lesssim n_d^{3/2} n^2 \cdot R,
\end{equation}
\begin{equation*}
    \begin{split}
        & \left\|\ba(T) - \ba(T-1) \right\|_2 \\
        = & \eta \cdot \left\| \frac{\partial \Tilde{\mathcal{L}}(T-1)}{\partial \ba}\right\|_2\\
        \leq & \eta \cdot \left\| \left[\begin{array}{cccccc}
            \frac{\partial s_1}{\partial \ba} & \cdots  & \frac{\partial s_{N_1}}{\partial \ba} & \frac{\partial \Bar{s}_{1}}{\partial \ba} & \cdots & \frac{\partial \Bar{s}_{N_2}}{\partial \ba}
        \end{array}\right]\right\|_2 \cdot \left\|\frac{\partial \Tilde{\mathcal{L}}(T-1)}{\partial \bm{s}} \right\|_2\\
        \leq & \eta \cdot \frac{N_1}{b_1} \sqrt{\sigma_{\max}} \sqrt{\mathcal{L}(T-1)} \\
        \lesssim & \eta \cdot \frac{N_1}{b_1} n_d^{3} n^{4} R,
    \end{split}
\end{equation*}
and
\begin{equation*}
    \eta \lesssim \frac{b_1}{N_1 n_d^3 n^4}.
\end{equation*}
It also guarantees that $\mathbb{E}\left[ \left\|\bm{\chi}(t) \right\|_2 \cdot \left\| \bs(t)\right\|_2\right] \leq \frac{\eta \sigma_{\min}}{8} \cdot \mathcal{L}(t)$, inspired by \Cref{appendix_proof_lemma_error_sgd}.
Moreover,
\begin{equation*}
    \begin{split}
        & \mathbb{E}\left[\left.\left\|\bs(t+1) - \bs(t) \right\|_2^2 \cdot \bm{1}_{T > t}\right|\mathcal{F}_{t}\right]\\
        \leq & \sigma_{\max} \cdot \mathbb{E}\left[\left.\left(\left\|\ba(t+1)-\ba(t)\right\|_2^2 + \left\|\bc(t+1)-\bc(t)\right\|_2^2\right) \cdot \bm{1}_{T > t}\right|\mathcal{F}_{t}\right]\\
        \lesssim & \frac{N_1}{b_1} \eta^2 n_d^{6} n^{8} \cdot \mathcal{L}(t),
    \end{split}
\end{equation*}
under the condition that
\begin{equation*}
    \eta \lesssim \frac{b_1 \sigma_{\min}}{N_1 n_d^6 n^8}.
\end{equation*}

We begin to show that the event $\mathcal{E} = \{T=\infty\}$ for physics-informed KANs with SGD holds with high probability. Following \Cref{lemma_event_never_leave_ball}, we have
\begin{equation*}
    \mathbb{E}\left[\sum_{t=0}^{T-1} \left\|\ba(t+1)-\ba(t) \right\|_2\right] 
        \lesssim \frac{n_d^{3/2} n^2}{\sigma_{\min}} \sqrt{\frac{N}{b}} \cdot \sqrt{\mathcal{L}(0)}.
\end{equation*}
Then, according to Markov inequality, we obtain
\begin{equation*}
\begin{split}
    & \mathbb{P}\left(\left\| \ba(T) -\ba(0)\right\|_2 > \frac{1}{2}R\right)\\
    \leq & \mathbb{P}\left(\sum_{t=0}^{T-1} \left\|\ba(t+1)-\ba(t) \right\|_2 > R\right) \\
    \leq & \frac{\mathbb{E}\left[\sum_{t=0}^{T-1} \left\|\ba(t+1)-\ba(t) \right\|_2\right]}{R} \\
    \lesssim & \frac{\frac{n_d^{3/2} n^2}{\sigma_{\min}} \sqrt{\frac{N}{b}} \cdot \sqrt{\mathcal{L}(0)}}{\frac{\sigma_{\min} \sqrt{m}}{n_d^{5/2} n^{7/2} \widetilde{M}_a^4 \widetilde{M}_c^2}},
\end{split}
\end{equation*}
which implies that
\begin{equation*}
    m \gtrsim \frac{N_1 n_d^{8} n^{11} \widetilde{M}_a^4 \widetilde{M}_c^2}{b_1 \sigma_{\min}^{4} \Tilde{\delta}^2}.
\end{equation*}

Moreover, for all $q \in [m]$, 
\begin{equation*}
   \begin{split}
        & \left\| \ba_{q}(T)\right\|_2 \\
        \leq & \left\| \ba_{q}(T) - \ba_{q}(0)\right\|_2 + \left\|\ba_{q}(0)\right\|_2\\
        \leq & \left\| \ba(T) - \ba(0)\right\|_2 + M_a.
   \end{split} 
\end{equation*}
Therefore,
\begin{equation*}
    \begin{split}
        & \mathbb{P}\left(\exists q \in [m], \left\| \ba_{q}(T)\right\|_2 > \frac{1}{2} \widetilde{M}_a \right)\\
        \leq & \mathbb{P}\left(\left\| \ba(T) - \ba(0)\right\|_2 +  M_a > \frac{1}{2} \widetilde{M}_a \right)\\
        \leq &  \mathbb{P}\left(\sum_{t=0}^{T-1} \left\|\ba(t+1)-\ba(0) \right\|_2 + M_a> \frac{1}{2} \widetilde{M}_a \right)\\
        \leq & \frac{\mathbb{E}\left[\sum_{t=0}^{T-1} \left\|\ba(t+1)-\ba(0) \right\|_2\right] + M_a}{\frac{1}{2} \widetilde{M}_a}.
    \end{split}
\end{equation*}
To guarantee that $\left\| \ba_{q}(T)\right\|_2 \leq \frac{1}{2} \widetilde{M}_a$ and $\left\| \bc_{q}(T)\right\|_2 \leq \frac{1}{2} \widetilde{M}_c$ with probability of at least $1-\Tilde{\delta}$, we equivalently require that
\begin{equation*}
    \widetilde{M}_a \simeq \frac{\frac{n_d^{3/2} n^2}{\sigma_{\min}} \sqrt{\frac{N}{b}} \cdot \sqrt{\mathcal{L}(0)} + M_a}{\Tilde{\delta}},
\end{equation*}
and
\begin{equation*}
    \widetilde{M}_c \simeq \frac{\frac{n_d^{3/2} n^2}{\sigma_{\min}} \sqrt{\frac{N}{b}} \cdot \sqrt{\mathcal{L}(0)} + M_c}{\Tilde{\delta}}.
\end{equation*}
In summary, we require that
\begin{equation*}
    \begin{split}
        m \gtrsim & \frac{N_1^7  n_d^{40} n^{63}}{b_1^7 \sigma_{\min}^{16} \Tilde{\delta}^{14}} \cdot \left(\log \frac{N_1 + N_2}{\delta}\right)^{7}\\
        & + \frac{N n_d^{10} n^{15}}{b \sigma_{\min}^4 \Tilde{\delta}^{14}} \cdot \left(n_d^6 n^4 + \left(\log \frac{m}{\delta}\right)^{6} \right) \cdot \left(\log \frac{N_1 + N_2}{\delta}\right).
    \end{split}
\end{equation*}

\end{document}